\documentclass[lettersize,journal]{IEEEtran}
\usepackage{amsmath,amsfonts}

\usepackage{algorithm}
\usepackage{array}
\usepackage[caption=false,font=normalsize,labelfont=sf,textfont=sf]{subfig}
\usepackage{textcomp}
\usepackage{stfloats}
\usepackage{url}
\usepackage{verbatim}
\usepackage{graphicx}
\usepackage{amssymb}
\usepackage{algpseudocode}
\usepackage{multirow}
\usepackage{booktabs}
\usepackage{enumitem}
\usepackage{cite}
\usepackage[colorlinks=true, citecolor=green, linkcolor=red]{hyperref}
\begin{document}
\title{Exploring Generalizable Distillation for Efficient Medical Image Segmentation}
\author{Xingqun Qi, Zhuojie Wu,  
Min Ren, Muyi Sun, \\
Caifeng Shan,~\IEEEmembership{Senior Member, IEEE}, and Zhenan Sun,~\IEEEmembership{Senior Member, IEEE}
\vspace{-13mm}
\thanks{This work was done when Xingqun Qi is a research intern at the CRIPAC, NLPR, Institute of Automation, Chinese Academy of Sciences, Beijing, China, 100190, (e-mail: xingqunqi@gmail.com).
This work is supported by China Postdoctoral Science Foundation under Grant No.2022M713362. (Corresponding author: Muyi Sun)}
\thanks{Zhuojie Wu is with the School of Artificial Intelligence /Automation, Beijing University of Posts and Telecommunications, Beijing, China, 100876, (e-mail:  zhuojiewu@bupt.edu.cn).}
\thanks{Caifeng Shan is with the College of Electrical Engineering and Automation, Shandong University of Science and Technology, Qingdao 266590, China, (e-mail: caifeng.shan@gmail.com).}
\thanks{Min Ren, Muyi Sun and Zhenan Sun are with the CRIPAC, NLPR, Institute of Automation, Chinese Academy of Sciences, Beijing, China, 100190, (e-mails: min.ren@cripac.ia.ac.cn; muyi.sun@cripac.ia.ac.cn; znsun@nlpr.ia.ac.cn).}}

\markboth{Journal of \LaTeX\ Class Files,~Vol.~14, No.~8, August~2021}%
{Shell \MakeLowercase{\textit{et al.}}: A Sample Article Using IEEEtran.cls for IEEE Journals}


\maketitle

\begin{abstract}
Efficient medical image segmentation aims to provide accurate pixel-wise predictions for medical images with a lightweight implementation framework. 
However, lightweight frameworks generally fail to achieve superior performance and suffer from poor generalizable ability on cross-domain tasks.
In this paper, we explore the generalizable knowledge distillation for the efficient segmentation of cross-domain medical images.
Considering the domain gaps between different medical datasets, we propose the Model-Specific Alignment Networks (MSAN) to obtain the domain-invariant representations.
Meanwhile, a customized Alignment Consistency Training (ACT) strategy is designed to promote the MSAN training. 
Considering the domain-invariant representative vectors in MSAN, we propose two generalizable knowledge distillation schemes for cross-domain distillation, Dual Contrastive Graph Distillation (DCGD) and Domain-Invariant Cross Distillation (DICD).
Specifically, in DCGD, two types of implicit contrastive graphs are designed to represent the intra-coupling and inter-coupling semantic correlations from the perspective of data distribution.
In DICD, the domain-invariant semantic vectors from the two models (\emph{i.e.}, teacher and student) are leveraged to cross-reconstruct features by the header exchange of MSAN, which achieves improvement in the generalization of both the encoder and decoder in the student model.
Furthermore, a metric named Fréchet Semantic Distance (FSD) is tailored to verify the effectiveness of the regularized domain-invariant features. 
Extensive experiments conducted on the Liver and Retinal Vessel Segmentation datasets demonstrate the superiority of our method, in terms of performance and generalization on lightweight frameworks. Our code will be available at \href{https://github.com/XingqunQi-lab/GKD-Framework}{\textit{GKD Framework}}
\end{abstract}


\section{Introduction}
\label{sec:introduction}
\IEEEPARstart{M}{EDICAL} image segmentation refers to obtaining the accurate pixel-level semantic prediction of clinical medical images. 
With its significant practical values for auxiliary diagnosis, medical image segmentation has impressive drawn progress with the development of deep learning \cite{b1,b2,b3}. 
Numerous works have emerged and demonstrated superior performances, even in complex scenarios (\emph{i.e.} cross-domain segmentation \cite{b4,b5,b6,b7,b8}). 
However, these models heavily rely on massive parameters and deeper network frameworks, which could not be applied on edge or embedded devices.
Meanwhile, due to the popularity of high-resolution most medical images (\emph{e.g.} $10^{5}\times10^{5}$ in \cite{b33}), shallow models are not sufficient to learn their rich spatial contextual information.
Obtaining high-accuracy segmentation from the medical images with lightweight models still remains challenging, especially in complex scenarios. 
To address this issue, various lightweight technologies have been taken into consideration for efficient medical image segmentation \cite{b10,b11}. 
\begin{figure}[t]
\begin{center}
\includegraphics[width=0.9\linewidth]{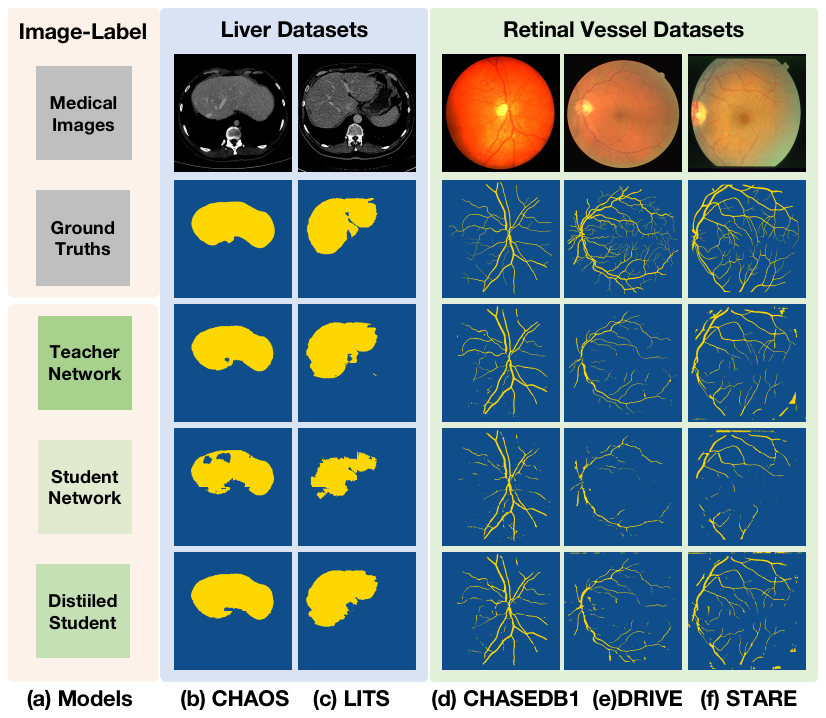}
\end{center}
\vspace{-3mm}
\caption{Liver and vessel segmentation results on cross-domain datasets of scratch-trained teacher, scratch-trained student, and the distilled student by our method. 
Segmented foreground pixels are colored in yellow. 
We test these samples on the liver datasets with the pre-trained model on the CHAOS training set and test the samples on vessel datasets with the pre-trained model on the CHASEDB1 training set.
Our generalizable knowledge distillation framework could improve the performance and generalization capability of the student networks jointly.}
\vspace{-5mm}
\label{fig:show_motivation}
\end{figure}

Lightweight technologies can be roughly divided into compression-based approaches \cite{b12,b14}, pruning-based approaches \cite{b15,b17}, and knowledge distillation-based ones \cite{b18,b19,b20,b21,b22}. 
Recently, knowledge distillation-based approaches have shown better potential for efficient medical image segmentation tasks compared with the other two types \cite{b12,b14,b15,b17}. 
Benefiting from the semantic information distilled from the powerful complex networks (called teacher), the lightweight models (called student) achieve significant performance improvement in single-domain tasks. 
However, the vast majority of these methods \cite{b23,b25} always overlook the poor performances of lightweight models on cross-domain datasets, even for the same semantic-class tasks as depicted in Fig. \ref{fig:show_motivation}. 
Generally, the complicated models have better domain-invariant representation capabilities which are not potently explored by previous knowledge distillation methods.
Therefore, it is worth exploring the high-performance knowledge distillation approaches with better generalization capability for efficient cross-domain medical image segmentation.

Transferring the domain-invariant features provides a good paradigm for the lightweight models to improve their generalizable abilities \cite{b26,b27,b28}. 
Among the approaches to extract the domain-invariant knowledge, contrastive learning based methods have illustrated superiority, due to the full utilization of enormous augmented data\cite{b29,b30,b31} through an unsupervised way. 
Therefore, in this paper, we intend to explore the generalizable knowledge by building contrastive correlations between the source images and their augmented samples. 
However, as for semantic segmentation, it is difficult to explicitly construct the pixel-wise contrastive supervision templates. 
Thus, we leverage two types of implicit contrastive graphs to represent the semantic domain-shift \cite{b52} between different augmented samples. Through distilling these graphs, the student model attains significant improvements on the cross-domain dataset.
Different from the previous works \cite{b32,b33}, the graphs involved in our framework provide more accessible implicit associations from teacher to student.

\begin{figure}[t]
\begin{center}
\includegraphics[width=0.9\linewidth]{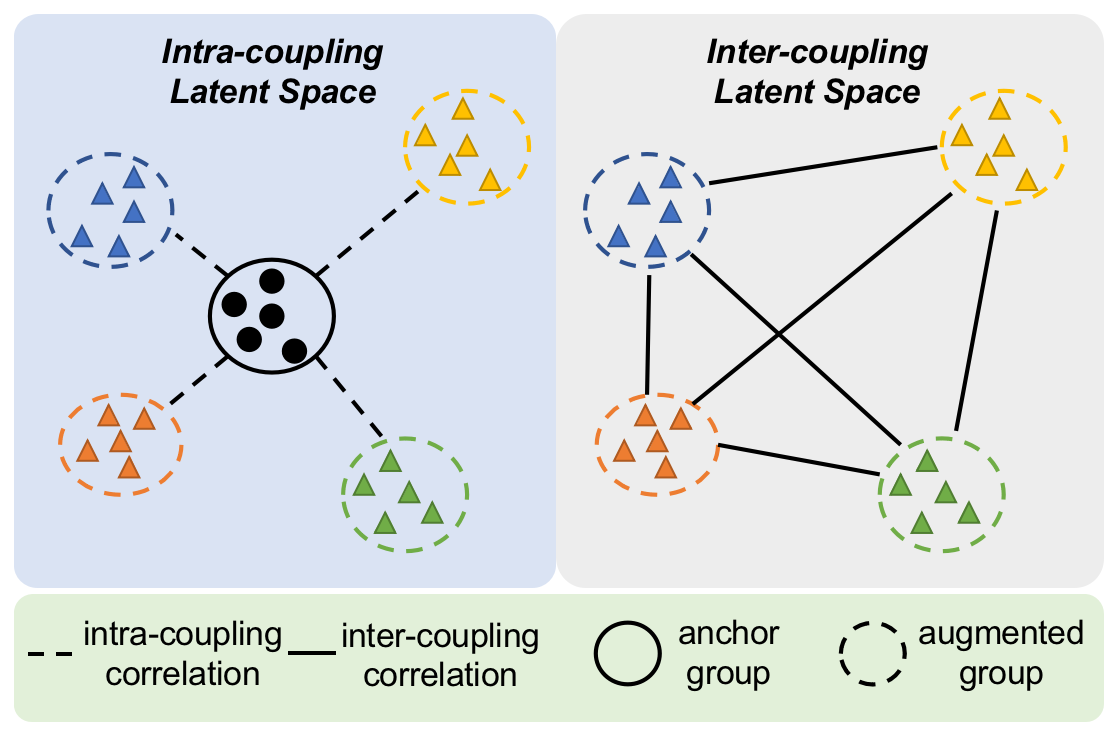}
\end{center}
\vspace{-5mm}
\caption{Algorithm structure of the proposed Generalizable Knowledge Distillation (GKD) framework.
The MSAN provides the semantic latent vectors.
Based on the MSAN, the pre-trained semantic autoencoder provides the domain-invariant regularization for the semantic vectors in MSAN. 
Then, the domain-invariant latent vectors are obtained and employed to design the generalizable distillation schemes (DCGD + DICD) for effective and generalizable distillation. }
\vspace{-5mm}
\label{fig:GKD}
\end{figure}

Motivated by the above discussions, we design a generalizable knowledge distillation framework (dubbed GKD) for efficient cross-domain medical image segmentation, as illustrated in Fig.~\ref{fig:GKD}.
The GKD framework can be treated as the mapping process from the domain-specific medical images to the domain-invariant dense predictions. 
In order to obtain consistent domain-invariant semantic representations from the teacher and student networks, we design the Model-Specific Alignment Networks (MSAN) with the Alignment Consistency Training (ACT) strategy.
Specifically, we pretrain a semantic autoencoder for obtaining the domain-invariant semantic representations. 
Then we leverage these domain-invariant representations to provide the supervision for training the Student Alignment Network (SAN) and Teacher Alignment Network (TAN) included by MSAN.

To enhance the performance and generalization ability of the student network, two generalizable knowledge distillation schemas are proposed, named Dual Contrastive Graph Distillation (DCGD) and Domain-Invariant Cross Distillation (DICD). 
Specifically, in DCGD, we construct two implicit contrastive graphs based on the domain-invariant semantic latent vectors from MSAN. 
These two types of graphs could demonstrate the intra-coupling and inter-coupling semantic correlations through the various data augmentation tactics, as depicted in Fig.~\ref{fig:intra+inter}. 
The distillation of these implicit `contrastive graphs' is more accessible to teacher-student distillation models compared with previous ones \cite{b29,b30,b51}. 

Additionally, we propose a DICD schema to constrain the encoder and decoder of the student model to be more adapted to the generalizable semantic knowledge. 
In particular, the domain-invariant latent vectors encoded from MSAN are leveraged to cross-reconstruct the features by exchanging headers of TAN and SAN. 
The reconstructed features from the student are fed into the decoder of the teacher to produce segmented predictions. 
The predictions are supervised by the original predictions of the teacher. 
A similar operation is also done for the reconstructed teacher features. 
In such a paradigm, the decoder of the student model is cross-trained with its encoder simultaneously in a consistent pattern. 

Furthermore, to verify the domain-invariant representative ability of our method, a new quantitative evaluation metric named Fréchet Semantic Distance (FSD) is tailored by the above-mentioned semantic autoencoder. 
We define the FSD as the Fréchet Distance between the latent features encoded by the student model (or teacher) and the semantic autoencoder, respectively. 
We demonstrate that the proposed metric is plausible in estimating the domain-invariant semantic knowledge. 

\noindent Overall, our contributions are summarized as follows:
\begin{itemize}[leftmargin=*]
\item A Generalizable Knowledge Distillation (GKD) framework is proposed to transfer the generalizable knowledge from the teacher to the student model, thus achieving efficient medical image segmentation in cross-domain scenarios.
\item The Model-Specific Alignment Networks (MSAN) incorporated with an Alignment Consistency Training (ACT) strategy are designed to obtain the domain-invariant representations, which facilitate generalizable distillation.
\item Two generalizable knowledge distillation schemes, named Dual Contrastive Graph Distillation (DCGD) and Domain-Invariant Cross Distillation (DICD) are proposed to boost the generalizable ability of the student model.
\item Extensive experiments on cross-domain medical image datasets verify that our method  outperforms the previous knowledge distillation counterparts significantly. 
\end{itemize}

\vspace{-3mm}
\section{Related Work}
\vspace{-1mm}
\subsection{Efficient Medical Image Segmentation} 
Medical image segmentation has developed rapidly in recent years due to its significant practical value. Excessive research based on convolutional neural networks has been proposed for this task. Most of the previous work has focused on architectural optimization based on the UNet framework \cite{b1}, which consists of the symmetric encoder and decoder. Valanarasu \emph{et al.} \cite{b34} proposed an over-complete architecture named Kiu-Net to map the deeper special features in blurred noisy medical image segmentation. 
Li \emph{et al.} \cite{b35} introduced a dual-direction attention block for accurate retinal vessel segmentation. 
Since the inevitable significant increase in computational cost caused by deepening and widening the network architectures, lightweight techniques have been introduced to remedy this issue \cite{b2, b37}. 
However, these lightweight frameworks are limited in solving cross-domain problems, though some works investigate the generalizable ability of the network in cross-domain medical image segmentation. 
For example, Jiang \emph{et al.} \cite{b4} applied a negative-transfer-resistant mechanism for cross-domain brain CT image segmentation. 
These works fail to reasonably explore the combination of lightweight techniques and generalization capabilities. 
In contrast, our method aims to jointly improve the performance of lightweight models and enhance their generalizable abilities.

\begin{figure}[t]
\begin{center}
\includegraphics[width=0.9\linewidth]{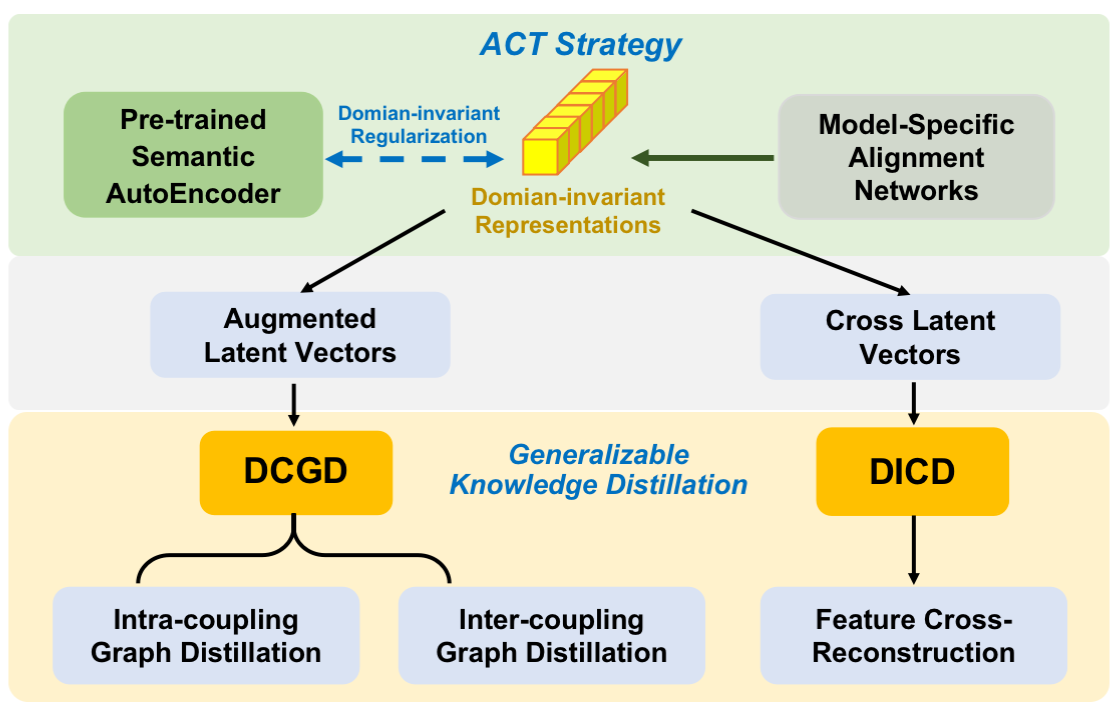}
\end{center}
\vspace{-5mm}
\caption{Explanations of the intra-coupling and inter-coupling correlations for graph distillation. 
Circles $\bigcirc$ represent anchor features. 
Triangles $\bigtriangleup$ represent augmented features. 
Different colors represent different data perturbation tactics. 
We try to calculate the intra-coupling correlations between the anchor features and each group of augmented features at the bottleneck level.
Meanwhile, as for augmented features, there are inter-coupling correlations among them.
Through this design, we aim to replace the previous single sample representation from the perspective of data distribution.}
\vspace{-5mm}
\label{fig:intra+inter}
\end{figure}

\vspace{-4.5mm}
\subsection{Knowledge Distillation}
Knowledge distillation provides the trade-off between network performance and computational cost. 
Conventionally, knowledge distillation aims to impart effective knowledge extracted from a complex teacher network to a shallow student network for improving the student's performance \cite{b18}. 
Zagoruyko \emph{et al.} \cite{b19} leveraged the intermediate-layer attention maps transferred from teacher to student, which increases the capability significantly. 
Yim \emph{et al.} \cite{b20} transferred knowledge flow between different layers from teacher to student, thus reaching better and faster optimization for the student.
Besides, Wang \emph{et al.} \cite{b22} extracted the intra-class feature variation to guide the lightweight model training for obtaining better performance. 
Zou \emph{at al.} \cite{b23} acquired the SOTA performance on pathological gastric cancer segmentation by enhancing the intermediate cross-layer correlation distilled from the teacher model. 
However, previous knowledge distillation methods only focus on the domain-specific task which bears poorly on cross-domain images. 
In contrast, we force the student to imitate the robust and generalizable knowledge from the teacher via implicit contrastive graphs.

\vspace{-3mm}

\subsection{Domain-invariant Representation}

Since conventional knowledge distillation methods fail to transfer generalizable information, the approaches which engage in exploring domain-invariant knowledge attract enormous attention in the range of cross-domain tasks. 
Li \emph{et al.} \cite{b28} extracted the mean difference of the inter-domain features as the transferable domain-invariant representations to enhance the generalizable ability. 
Creager \emph{et al.} \cite{b38} designed a general framework for domain-invariant learning to maximize information for downstream invariant learning.
Moreover, some researchers introduce contrastive learning based approaches to extract domain-invariant knowledge \cite{b30, b31}. 
He \emph{et al.} \cite{b32} built a contrastive dynamic dictionary as domain-invariant features which facilitate the unsupervised vision tasks. 
The existing methods explicitly construct image-level positive-negative contrastive samples, thus failing to be directly applied to segmentation tasks. In contrast, the implicit contrastive graphs involved in our method represent more accessible generalizable knowledge for the student in pixel-wise prediction.

\begin{figure}[t]
\begin{center}
\includegraphics[width=0.9\linewidth]{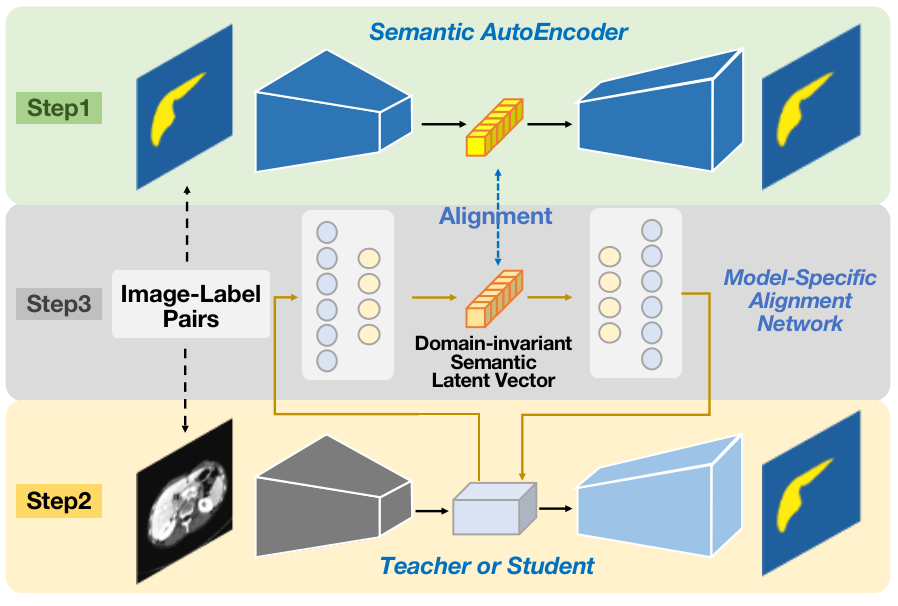}
\end{center}
\vspace{-5mm}
\caption{Pipeline of our proposed Alignment Consistency Training (ACT) strategy for the Model-Specific Alignment Network (MSAN). 
Step 1: Pre-train the Semantic AutoEncoder; 
Step 2: Train the teacher and student model respectively, only with task-driven supervision; 
Step 3: Train the Model-Specific Alignment Networks (TAN and SAN) of pre-trained teacher and pre-trained student models, respectively.
Then the Domain-invariant Semantic Latent Vectors of each model could be obtained.}
\vspace{-5mm}
\label{fig:ACT}
\end{figure}

\section{Proposed Method}
\vspace{-1mm}
The proposed GKD framework aims at exploring the generalizable distillation knowledge, extracted from a powerful and complicated teacher model $T$, to improve the performance of the lightweight student model $S$ on the medical image segmentation. 
The student model distilled by GKD achieves upgraded results on the other domain images. Our GKD framework consists of three components: the Model-Specific Alignment Networks (MSAN), the Dual Contrastive Graph Distillation (DCGD) schema, and the Domain-Invariant Cross Distillation (DICD) schema.
 
\vspace{-3mm}
\subsection{Model-Specific Alignment Networks}
Model-Specific Alignment Networks consist of two alignment networks, Student Alignment Network (SAN) and Teacher Alignment Network (TAN), which are designed for the extraction of domain-invariant semantic latent vectors. 
As depicted in Fig.~\ref{fig:ACT}, our proposed MSAN is guided by the Alignment Consistency Training (ACT) strategy. 
By leveraging the ACT, a semantic autoencoder is first pre-trained to provide the domain-invariant regularization for the above semantic latent vectors extracted by MSAN. 
In this way, the encoded semantic features of the teacher and student could be mapped to the regularized latent representations by MASN. 
Specifically, the details of the ACT strategy are presented below.

\begin{figure}[t]
\begin{center}
\includegraphics[width=0.92\linewidth]{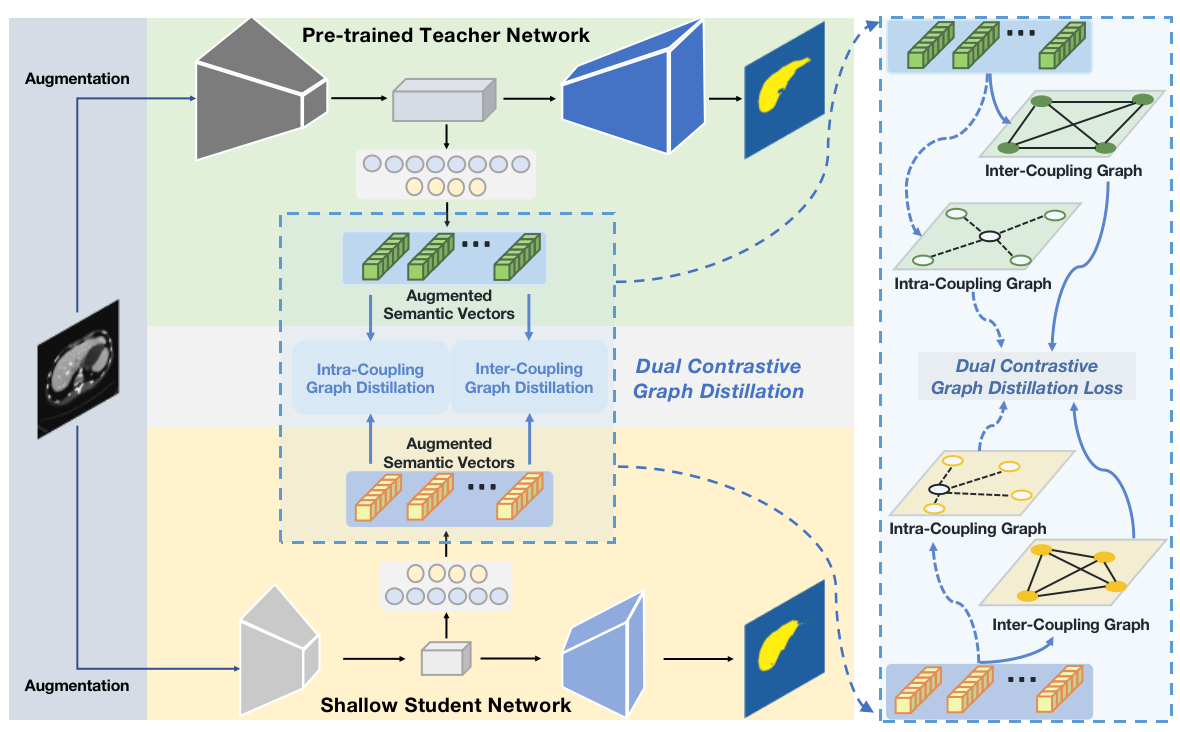}
\end{center}
\vspace{-5mm}
\caption{The proposed Dual Contrastive Graph Distillation (DCGD) schema (left) and the detailed design in DCGD (right). 
Each training sample is fed into two models simultaneously after different data augmentation strategies jointly. The MSAN encoder is exploited to extract domain-invariant knowledge.
We use the anchor vector and the augmented semantic vectors to construct two types of contrastive graphs for distillation. 
The pre-trained teacher network and MSAN structure are frozen when the student model is distilled during training.}
\vspace{-6.5mm}
\label{fig:DCGD}
\end{figure}

\subsubsection{Step 1: Pre-train a Semantic Autoencoder}
Initially, given a domain-specific medical image $X$ and a paired domain-invariant semantic annotation $Y$, 
the mapping from $X$ to $Y$ heavily relies on the semantic-aware features, which are extracted by the encoders of teacher and student. 
Thus, we pretrain a semantic autoencoder to explicitly provide the domain-invariant representations upon the semantic-aware features. 
Formally, the annotation is embedded as the latent vector $S^{Y}\in \mathbb{R}^{1\times C}$ (the yellow vector in Fig.~\ref{fig:ACT}, Step 1), where $C$ represents the vector dimension. 
The semantic autoencoder is supervised by the reconstruction loss, expressed as $\mathcal{L}_{srec}=\left \| Y - \hat{Y}\right \|_{1}$,
where $\hat{Y}$ indicates the reconstructed annotation.
The embedded latent vector is leveraged to provide the domain-invariant regularization in the following step 3.

\subsubsection{Step 2: Obtain the Scratch-trained Teacher and Student networks} 
We find that in previous knowledge distillation research \cite{b18,b22}, simply transferring the prediction from the teacher to the student is too ragged for student learning. 
Despite other works intending to distill the intermediate-layer knowledge \cite{b19,b20,b21}, the deep-level semantic information in the bottleneck block after the teacher encoder would be neglected easily. 
Thus, we transfer this bottleneck-level semantic information to the shallow student network. 
In step 2 of ACT, we provide the conventional semantic segmentation supervision for the trained-from-scratch teacher and student models. 
This stage engages in obtaining the encoded bottleneck-level semantic features from the teacher and student. 
Then the parameters of the teacher and student models are fixed in the next step.
The teacher and student are trained by the basic cross-entropy loss:
\begin{align}
\mathcal{L}_{ce}=-\frac{1}{N} \sum ( Y -\log_{}{(G^{T/S}(X))) } ,
\end{align}
where $G^{T/S}$ indicates the teacher or student models.

\subsubsection{Step 3: Obtain the Model-Specific Alignment Networks} 
Driven by the aforementioned two steps, we develop the Model-Specific Alignment Networks (MSAN) for obtaining domain-invariant semantic latent vectors, as shown in Fig.~\ref{fig:ACT} \emph{(step 3)}. 
Structurally, the MSAN (TAN or SAN) consists of a pair of symmetrical encoding-header and decoding-header.
Since the bottleneck-level semantic features $F^{T/S}$
of teacher and student are different with the spatial dimensions, 
we utilize the encoding header to map these features into the latent vectors $S^{T/S} \in \mathbb{R}^{1\times C}$ with a unified dimension $C$.
\textbf{Importantly}, we exploit the corresponding latent vectors acquired by the pre-trained semantic autoencoder for domain-invariant regularization, expressed as:
\begin{align}
\mathcal{L}_{regu}=\left \|S^{Y}-S^{T/S}  \right \|_{1}. 
\end{align}
Then, the semantic vectors $S^{T/S}$ are upsampled into feature maps by the model-specific decoding headers, which could also be treated as the feature map reconstruction to minimize the gap between the input and output of MSAN. This process is supervised by the following loss:
\begin{align}
\mathcal{L}_{rec}^{F^{T/S}} = \left \| F^{T/S}-\hat{F}^{T/S}  \right \|_{1},
\label{eq3}
\end{align}
where the $\hat{F}^{T/S}$ denotes reconstructed features. 
In this step, the overall MSAN is optimized by:
\begin{align}
\mathcal{L}_{MSAN}= \mathcal{L}_{rec}^{F^{T/S} } + \lambda  \mathcal{L}_{regu},
\label{eq4}
\end{align}
where $\lambda$ is the loss weight term which is set 0.5 in practice.
Supervised by Eq. \eqref{eq4}, the encoded bottleneck-level semantic features from the teacher and student models are mapped to a consistent domain-invariant latent space.
After the training of MSAN, the encoding and decoding headers are fixed, and only the student model is optimized in the following Dual Contrastive Graph Distillation (DCGD) and Domain-Invariant Cross Distillation (DICD) schemes.

\vspace{-3mm}
\subsection{Dual Contrastive Graph Distillation}
Generally, in medical image segmentation, the deeper network plays an important role in extracting sufficient semantic information. 
Therefore, shallow student networks usually perform less successfully compared to complicated teacher networks, especially on cross-domain datasets. 
To transfer this precious knowledge from teacher to student, an intuitive manner is to enforce the spatial and dimensional feature alignment between the teacher and student. 
However, rough feature alignment (\emph{e.g.} interpolation between feature maps of different spatial sizes) could lead to semantic information missing, and hence the performance of the student model is limited \cite{b19,b23}. 
Meanwhile, the generalizable ability of student models is often overlooked.

To this end, we jointly take these two factors into account for improving the performance and generalization capability of the student networks.
Benefiting from the predesignated MSAN, the bottleneck-level features of these two models are aligned in a consistent domain-invariant pattern. 
To further investigate the generalizable knowledge representation, we propose a Dual Contrastive Graph Distillation (DCGD) schema. 
It is worth noting that in DCGD, the contrastive graphs implicitly introduce more knowledge about the data distribution than the representation of a single sample, which could bring more generalization and robustness. 

Specifically, the input medical image is augmented by various tricks (\emph{i.e.} Random Cutout, Sobel Filter, Gaussian noise, Gaussian blur), expressed as: $X\longmapsto \mathbb{X} =\left \{ \hat{x}, x_{1},x_{2},\dots ,x_{i}\right \} $, where $\hat{x}$ represents the original anchor sample and $x_{i}$ denotes $i$ $th$ augmented sample. For the anchor sample and each augmented sample, there are different contrastive couplings, expressed as $\mathcal{C} = \left \{ (\hat{x}, x_{1}), (\hat{x}, x_{2} ),\cdots, (\hat{x}, x_{i} )\right \}$.
We feed the $\mathbb{X}$ into both models (\emph{i.e.} teacher and student) simultaneously and construct two implicit contrastive graphs to represent the generalizable knowledge. 
As shown in Fig.~\ref{fig:DCGD}, our DCGD schema consists of two sub-distillation methods: (i) the intra-coupling Contrastive Graph Distillation and (ii) the inter-coupling Contrastive Graph Distillation.

\begin{figure}[t]
\begin{center}
\includegraphics[width=0.92\linewidth]{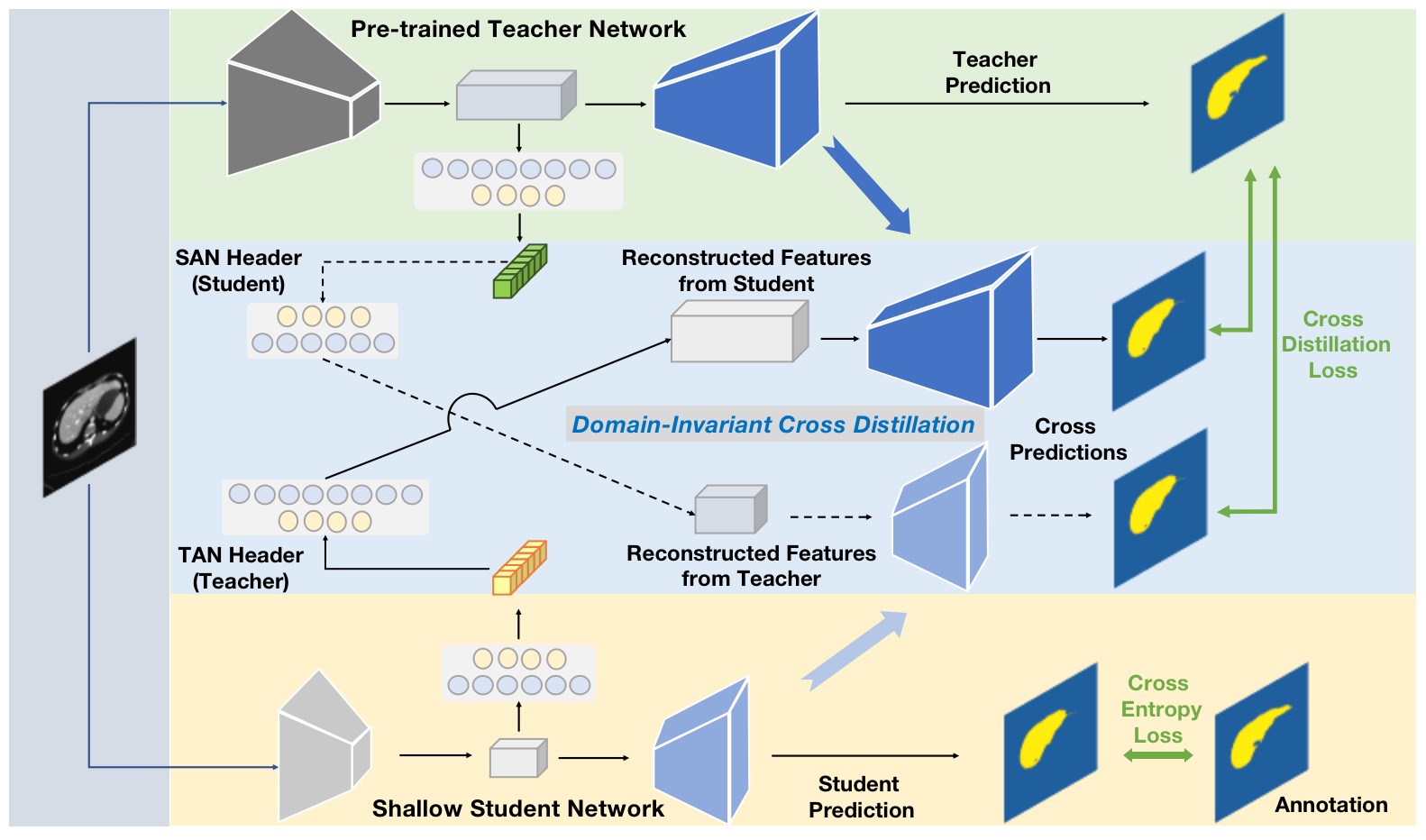}
\vspace{-5mm}
\end{center}\caption{The pipeline of our proposed Domain-Invariant Cross Distillation (DICD) schema. In DICD, the model-specific features from teacher and student are cross-reconstructed by the header exchanging of MSAN. The pre-trained teacher model is utilized to provide the cross distillation supervision on both the encoder and decoder of the student respectively.}
\vspace{-6mm}
\label{fig:DICD}
\end{figure}

\subsubsection{Intra-coupling Contrastive Graph Distillation}
With the various data augmentation strategies, the teacher and student models both receive $\mathbb{X}$ as the inputs. 
Intuitively, for each anchor sample, the different data augmentations result in different distribution shifts of the semantic margin on the domain-specific medical images, as shown in Fig.~\ref{fig:intra+inter}.
The encoder of the teacher networks could obtain more robust and generalizable representations with different augmented samples. 
Since the semantic annotations of augmented samples are difficult to obtain for supervision, we construct the \textbf{implicit contrastive graphs} in the semantic latent space for bottleneck-level distillation. 

These implicit contrastive graphs represent the semantic-shift \cite{b52} between the anchor sample and various augmented samples in the different couplings.
Thus, the intra-coupling contrastive graph is formulated as $\mathcal{G}_{intra} < \mathcal{N}, \mathcal{E}>$, where $\mathcal{N}$ indicates the nodes in the graph; $\mathcal{E}$ denotes the edges between the anchor node and the augmented nodes in different couplings. The anchor node and each augmented node are obtained by MSAN encoding headers of the teacher and student, respectively. 
Specifically, the anchor sample and each augmented sample in different couplings are fed into the student and fixed teacher encoders to obtain the model-specific bottleneck-level features. 
Then, we leverage the fixed MASN encoding headers to embed the model-specific features into the domain-invariant semantic latent vectors: $S_{\mathbb{X}}^{T/S}\in \mathbb{R}^{1\times C}$, where $C = 512$ in practice.

In the intra-coupling contrastive graph, we intend to transfer the intra-coupling semantic-shift of the teacher model caused by data perturbation. 
Thus, we transpose the $S_{\mathbb{X}}^{T/S}$ and employ the matrix multiplications to obtain the intra-coupling similarity maps as the graph nodes $\mathcal{N}^{T/S}\in \mathbb{R}^{C\times C}$. As illustrated in Fig.~\ref{fig:DCGD}, we calculate the cosine similarity between the anchor node and each augmented node in different couplings as the edges $\mathcal{E}^{T/S}$ in the intra-coupling contrastive graph.
Finally, we exploit the $L_{1}$ loss to transfer this generalizable knowledge from teacher to student:
\begin{align}
\mathcal{L}_{intra} = &\mathcal{L}_{intra}(\mathcal{N}) +\mathcal{L}_{intra}(\mathcal{E}) 
\notag
\\&\left \| \mathcal{N}^{T}-\mathcal{N}^{S} \right \|_{1} + \left \| \mathcal{E}^{T}-\mathcal{E}^{S} \right \|_{1}.
\label{eq5}
\end{align}
In Eq. \eqref{eq5}, the implicit semantic intra-coupling contrastive graph distillation is expressed. 

\subsubsection{Inter-coupling Contrastive Graph Distillation}
Furthermore, we propose the inter-coupling contrastive graph $\mathcal{G}_{inter} < \mathcal{N}, \mathcal{E}>$ to represent the inter-coupling semantic-shift among each node as illustrated in Fig.~\ref{fig:intra+inter}. We argue that as for different augmented samples, there are inter-coupling semantic-shift caused by different data perturbations, which bring more robust and generalizable information from the perspective of data distribution. 
The inter-coupling contrastive graph aims to formulate the correlations between different augmented nodes as generalizable knowledge which is transferred to the student. Thus, in $\mathcal{G}_{inter}$, all the nodes are encoded from augmented samples with different couplings as shown in Fig.~\ref{fig:DCGD}. All the inter-graph edges are calculated among each node as cosine-similarity. 
Different from intra-coupling graph distillation, we regularize each node in inter-graph with the softmax function and then conduct the Kullback-Leibler Divergence, expressed as: 
\begin{align}
\mathcal{L}_{inter}(\mathcal{N}) & = KL(soft(\mathcal{N}^{T})\left |  \right | soft(\mathcal{N}^{S})),
\end{align}
where $soft(\cdot )$ indicates the softmax operation. Meanwhile, we utilize the $L_{1}$ loss to constraint with paired inter-graph edges.
\begin{align}
\mathcal{L}_{inter}(\mathcal{E})=\left \| \mathcal{E}^{T}-\mathcal{E}^{S} \right \|_{1}.
\end{align}
In summary, the inter-coupling contrastive graph distillation is formulated as:
\begin{align}
\mathcal{L}_{inter} = \mathcal{L}_{inter}(\mathcal{N}) +\mathcal{L}_{inter}(\mathcal{E}).        
\end{align}

\begin{table*}[t]
\centering
\caption{Comparison results on CHAOS and LITS datasets. All the models are only trained on the CHASO dataset, and directly tested on both datasets. $\uparrow$ indicates the higher, the better. $\downarrow$ indicates the lower, the better.}
\label{tab:my-table1}
\footnotesize
\resizebox{\textwidth}{!}{%
\renewcommand{\arraystretch}{1.0} 
\begin{tabular}{c|ccc|ccc|ccc|ccc|ccc}
\toprule
\textbf{Model}   & \multicolumn{3}{c|}{\textbf{SE $\uparrow$}}                                         & \multicolumn{3}{c|}{\textbf{ACC $\uparrow$}}                                        & \multicolumn{3}{c|}{\textbf{AUC $\uparrow$}}                                        & \multicolumn{3}{c|}{\textbf{F1 $\uparrow$}}                                         & \multicolumn{3}{c}{\textbf{mIOU $\uparrow$}}                                        \\ \hline
Dataset          & CHAOS           & \multicolumn{1}{c|}{LITS}            & \textbf{GAP$\downarrow$}    & CHAOS           & \multicolumn{1}{c|}{LITS}            & \textbf{GAP$\downarrow$}    & CHAOS           & \multicolumn{1}{c|}{LITS}            & \textbf{GAP$\downarrow$}    & CHAOS           & \multicolumn{1}{c|}{LITS}            & \textbf{GAP$\downarrow$}    & CHAOS           & \multicolumn{1}{c|}{LITS}            & \textbf{GAP$\downarrow$}    \\ \hline
\textbf{T:FANet} & \textbf{0.9613} & \multicolumn{1}{c|}{\textbf{0.8356}} & \textbf{0.1257} & \textbf{0.9932} & \multicolumn{1}{c|}{\textbf{0.9835}} & \textbf{0.0097} & \textbf{0.9986} & \multicolumn{1}{c|}{\textbf{0.9908}} & \textbf{0.0078} & \textbf{0.9401} & \multicolumn{1}{c|}{\textbf{0.8647}} & \textbf{0.0754} & \textbf{0.9399} & \multicolumn{1}{c|}{\textbf{0.8721}} & \textbf{0.0678} \\
S: Mobile Unet   & 0.8840          & \multicolumn{1}{c|}{0.4812}          & 0.4028          & 0.9899          & \multicolumn{1}{c|}{0.9625}          & 0.0274          & 0.9972          & \multicolumn{1}{c|}{0.9723}          & 0.0249          & 0.9057          & \multicolumn{1}{c|}{0.6180}          & 0.2877          & 0.9085          & \multicolumn{1}{c|}{0.7042}          & 0.2043          \\ \hline
KD               & 0.9349          & \multicolumn{1}{c|}{0.7296}          & 0.2053          & 0.9883          & \multicolumn{1}{c|}{0.9734}          & 0.0149          & 0.9971          & \multicolumn{1}{c|}{0.9847}          & 0.0124          & 0.8979          & \multicolumn{1}{c|}{0.7755}          & 0.1224          & 0.9012          & \multicolumn{1}{c|}{0.8027}          & 0.0985          \\
AT               & 0.9394          & \multicolumn{1}{c|}{0.7212}          & 0.2182          & 0.9893          & \multicolumn{1}{c|}{0.9737}          & 0.0156          & 0.9973          & \multicolumn{1}{c|}{0.9805}          & 0.0168          & 0.9062          & \multicolumn{1}{c|}{0.7757}          & 0.1305          & 0.9086          & \multicolumn{1}{c|}{0.8030}          & 0.1056          \\
FSP              & 0.9320          & \multicolumn{1}{c|}{0.7140}          & 0.2180          & 0.9912          & \multicolumn{1}{c|}{0.9740}          & 0.0172          & 0.9973          & \multicolumn{1}{c|}{0.9873}          & 0.0100          & 0.9209          & \multicolumn{1}{c|}{0.7763}          & 0.1446          & 0.9220          & \multicolumn{1}{c|}{0.8036}          & 0.1184          \\
SKD              & 0.9363          & \multicolumn{1}{c|}{0.7983}          & 0.1380          & 0.9901          & \multicolumn{1}{c|}{0.9764}          & 0.0137          & 0.9977          & \multicolumn{1}{c|}{0.9863}          & 0.0114          & 0.9121          & \multicolumn{1}{c|}{0.8103}          & 0.1018          & 0.9139          & \multicolumn{1}{c|}{0.8281}          & 0.0858          \\
IFVD             & 0.9250          & \multicolumn{1}{c|}{0.7175}          & 0.3075          & 0.9907          & \multicolumn{1}{c|}{0.9752}          & 0.0155          & 0.9977          & \multicolumn{1}{c|}{0.9843}          & 0.0134          & 0.9164          & \multicolumn{1}{c|}{0.7851}          & 0.1313          & 0.9180          & \multicolumn{1}{c|}{0.8101}          & 0.1079          \\
CoCo             & 0.9255          & \multicolumn{1}{c|}{0.7292}          & 0.1933          & 0.9917          & \multicolumn{1}{c|}{0.9764}          & 0.0153          & 0.9980          & \multicolumn{1}{c|}{0.9783}          & 0.0197          & 0.9251          & \multicolumn{1}{c|}{0.7960}          & 0.1291          & 0.9260          & \multicolumn{1}{c|}{0.8182}          & 0.1078          \\ \hline
\textbf{GKD}     & \textbf{0.9451} & \multicolumn{1}{c|}{\textbf{0.8251}} & \textbf{0.1200} & \textbf{0.9924} & \multicolumn{1}{c|}{\textbf{0.9825}} & \textbf{0.0099} & \textbf{0.9984} & \multicolumn{1}{c|}{\textbf{0.9912}} & \textbf{0.0072} & \textbf{0.9320} & \multicolumn{1}{c|}{\textbf{0.8560}} & \textbf{0.0760} & \textbf{0.9323} & \multicolumn{1}{c|}{\textbf{0.8649}} & \textbf{0.0674} \\ \bottomrule
\end{tabular}%
}
\vspace{-5mm}
\end{table*}

\vspace{-5mm}
\subsection{Domain-Invariant Cross Distillation}
To further improve the performance of the student model, we propose a Domain-Invariant Cross Distillation (DICD) schema based on the aforementioned MSAN as shown in Fig.~\ref{fig:DICD}. 
We notice that a powerful decoder which can be adaptive with generalizable encoded features also plays a crucial role in cross-domain medical image segmentation. 
To this end, we enhance the encoder and decoder of the student model jointly in a cross-training manner. 

Concretely, we leverage the cross-reconstructed bottleneck-level features to impose the consistency constraint on the network predictions.
Guided by MSAN, we obtain the regularized domain-invariant semantic latent vectors of anchor samples encoded by the teacher and student, respectively. 
Since the bottleneck-level feature dimensions of these two networks (\emph{i.e.} teacher and student) are different, we utilize the pre-trained MSAN decoding headers to obtain the cross-reconstructed features. 
Afterward, the cross-reconstructed features are fed into the exchanged network decoders to produce cross-predictions, respectively. 
Then, we exploit the original teacher prediction $Y^{T}$ to provide the $L_{1}$ constraint on the cross-predictions, separately. 
As expressed below:
\begin{align}
\mathcal{L}_{DICD} & = \left \| Y^{T} -\hat{Y}_{rec}^{T} \right \| _{} + \left \| Y^{T} -\hat{Y}_{rec}^{S} \right \| _{},
\end{align}
where $\hat{Y}_{rec}^{T}$ denotes the cross-prediction from teacher decoder, $\hat{Y}_{rec}^{S}$ denotes the cross-prediction from student decoder.
With the proposed DICD, the student model receives cross-reconstructed features from the teacher to further improve the segmentation performance and pursue more generalizable ability. 
Meanwhile, in DICD, the student model is also supervised by the basic cross-entropy loss $\mathcal{L}_{ce}$. 
\vspace{-3mm}
\subsection{Objective Functions}
The optimization of our proposed generalizable knowledge distillation (GKD) framework is divided into 3 phases. 
Firstly, the teacher and student models are trained by basic cross-entropy loss, while the parameters of the teacher model are frozen in the following phases. 
Then, we employ the $\mathcal{L}_{MSAN}$ to train the MSAN of both two models and only the parameters of MSAN are updated in this phase. 
In the last phase, we aim at obtaining the generalizable student model based on the aforementioned pre-trained teacher model and MSAN. The total training loss $\mathcal{L}_{total}$ is expressed as:
\begin{align}
\mathcal{L}_{total} = \mathcal{L}_{ce} + \alpha \mathcal{L}_{intra} + \beta \mathcal{L}_{inter} + \gamma \mathcal{L}_{DICD},
\end{align}
where the hyper-parameters $\alpha$, $\beta$ and $\gamma$ are set as 100, 100, 0.5 in practice, respectively.

\section{Experiments}

\begin{figure}[t]
\begin{center}
\includegraphics[width=0.9\linewidth]{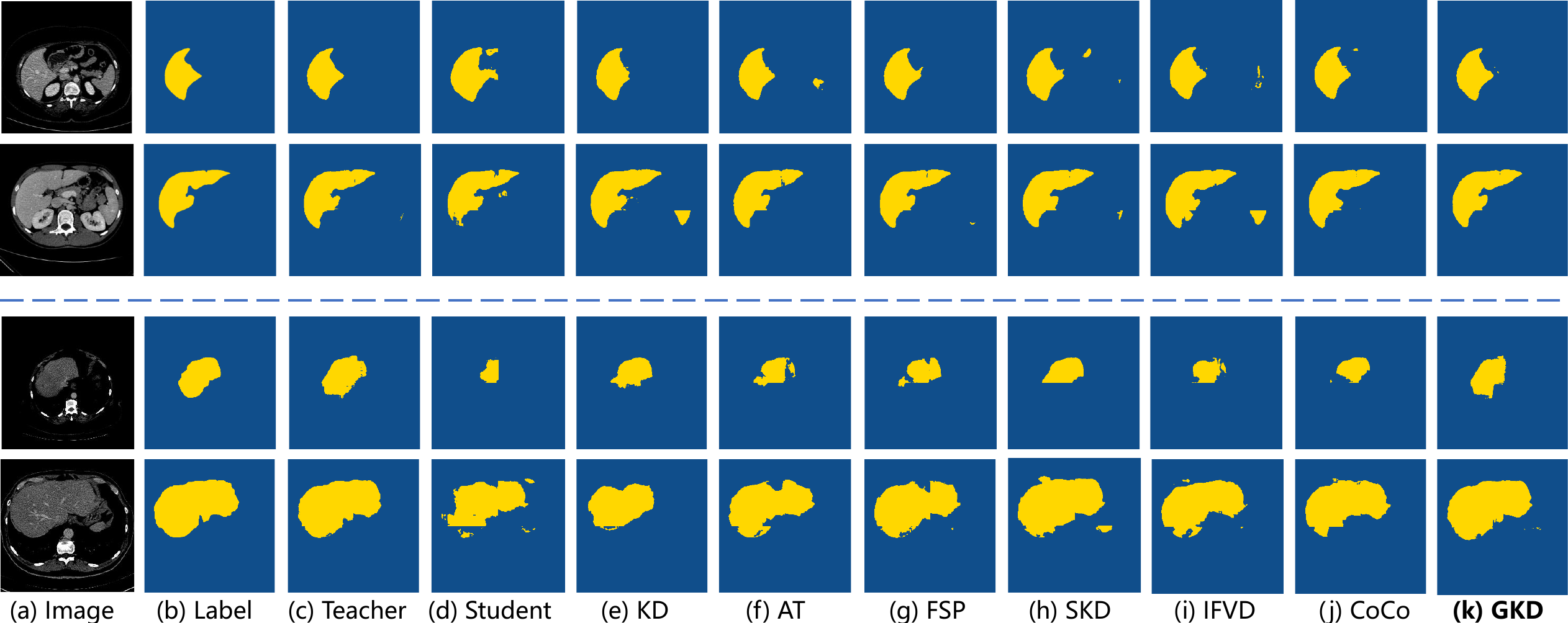}
\vspace{-4mm}
\end{center}\caption{Liver segmentation comparisons of the proposed GKD with state-of-the-art knowledge distillation counterparts. The top two-row samples are from the CHAOS dataset, and the bottom two-row samples are from the LITS dataset. \textbf{Zoom in for better details.}}
\vspace{-6mm}
\label{fig:liver_results}
\end{figure}

\subsection{Datasets}

\subsubsection{Liver Segmentation Datasets}
We conduct extensive experiments on the CHAOS \cite{b40} and LITS \cite{b41} liver segmentation datasets as depicted in Fig.~\ref{fig:show_motivation}. 
We focus on the CT modality images in the CHAOS liver segmentation dataset. 
CHAOS contains CT liver images of 40 patients whose livers are \textbf{healthy and not diseased}. 
In the original CHAOS dataset, both the training and testing sets contain images of 20 patients. 
However, only the original training set provides binary-class ground truth. 
We choose the images of 13 patients (\emph{i.e.} patient number: 1, 2, 5, 6, 8, 10, 14, 16, 18, 19, 21, 22, 23) from the original training set as the sub-training set for this experiment, and the remaining 7 patient images (\emph{i.e.} patient number: 24, 25, 26, 27, 28, 29, 30) are used as the sub-testing set.
Thus, we obtain 1,578 CT liver images as the sub-training set and 1,296 CT liver images as the sub-testing set. 
Each CT image in CHAOS has a size of $512 \times 512$ pixels. 
Moreover, we introduce the LITS liver segmentation dataset to verify the generalizable ability of the distilled student by our proposed GKD. 
The original LITS dataset contains numerous CT liver images of \textbf{diseased patients}, which is challenging due to the uneven shape and diffuseness of the images. 
Randomly, we pick 1,000 liver images with ground truth from LITS and merge their semantic categories into two categories: liver and background. 
The selected liver images are leveraged as a generalization test set which is not involved in the training stage. 
All the images in LITS have a size of $512 \times 512$ pixels.
\subsubsection{Retinal Vessel Segmentation Datasets}
We also perform experiments on three retinal vessel segmentation datasets CHASEDB1 \cite{b42}, STARE \cite{b43} and DRIVE \cite{b44}. 
The CHASEDB1 dataset consists of 24 retinal images with a spatial size of $999 \times 960$. 
The images are from the left and right eyes of 14 children. 
We select the retinal vessel images of the first 7 children as the training set and the remaining images as the testing set. 
The STARE dataset contains 20 retinal vessel images with a spatial size of $700 \times 604$. Similar to \cite{b35}, we directly evaluate the generalizable ability of the student model on the testing set of the STARE.
The DRIVE dataset includes 40 images of retinal vessels, each with a spatial size of $565 \times 584$. Similarly, we employ the 20 testing images of the DRIVE dataset as the generalization testing set to verify our proposed GKD framework. 

Notably, we only involve CHAOS and CHASEDB1 datasets in the training phase to obtain the student model and test directly on the other datasets. 
Besides, all the images are cropped into $128 \times 128$ size patches as the model input.

\begin{table*}[t]
\centering
\caption{Results on CHASEDB1, STARE and DRIVE datasets. All the models are only trained on the CHASEDB1 dataset, and directly tested on three retinal vessel datasets. $\uparrow$ indicates the higher, the better.}
\label{tab:my-table2}
\footnotesize
\resizebox{\textwidth}{!}{%
\renewcommand{\arraystretch}{1.0} 
\begin{tabular}{c|ccc|ccc|ccc|ccc|ccc}
\toprule
\textbf{Model}   & \multicolumn{3}{c|}{\textbf{SE $\uparrow$}}                    & \multicolumn{3}{c|}{\textbf{ACC $\uparrow$}}                   & \multicolumn{3}{c|}{\textbf{AUC $\uparrow$}}                   & \multicolumn{3}{c|}{\textbf{F1 $\uparrow$}}                    & \multicolumn{3}{c}{\textbf{mIOU $\uparrow$}}                   \\ \hline
Dataset          & CHASEDB1        & STARE           & DRIVE           & CHASEDB1        & STARE           & DRIVE           & CHASEDB1        & STARE           & DRIVE           & CHASEDB1        & STARE           & DRIVE           & CHASEDB1        & STARE           & DRIVE           \\ \hline
\textbf{T:FANet} & \textbf{0.8103} & \textbf{0.6485} & \textbf{0.5444} & \textbf{0.9730} & \textbf{0.9544} & \textbf{0.9597} & \textbf{0.9836} & \textbf{0.9541} & \textbf{0.9561} & \textbf{0.7711} & \textbf{0.6526} & \textbf{0.6558} & \textbf{0.7996} & \textbf{0.7183} & \textbf{0.7229} \\
S: Mobile Unet   & \textcolor{blue}{0.6894}          & \textcolor{blue}{0.5050}          & \textcolor{blue}{0.4174}          & 0.9686          & 0.9575          & 0.9543          & 0.9737          & 0.9244          & 0.9320          & 0.7113          & 0.6106          & 0.5630          & 0.7596          & 0.6977          & 0.6723          \\ \hline
KD               & 0.6921          & 0.4958          & 0.3855          & 0.9715          & 0.9608 & 0.9538          & 0.9800          & 0.9425          & 0.9450          & 0.7315          & 0.6254          & 0.5410          & 0.7735          & 0.7072          & 0.6617          \\
AT               & 0.7150          & 0.5391          & 0.4277          & 0.9716          & 0.9598          & 0.9551          & 0.9788          & 0.9391          & 0.9377          & 0.7385          & 0.6391          & 0.5732          & 0.7779          & 0.7139          & 0.6777          \\
FSP              & 0.7345          & 0.5193          & 0.4125          & 0.9698          & 0.9587          & 0.9548          & 0.9770          & 0.9213          & 0.9258          & 0.7320          & 0.6242          & 0.5631          & 0.7729          & 0.7055          & 0.6727          \\
SKD              & 0.7478          & 0.5677          & 0.4515          & 0.9701          & 0.9584          & 0.9570          & 0.9792          & 0.9300          & 0.9106          & 0.7369          & 0.6434          & 0.5972          & 0.7761          & 0.7155          & 0.6907          \\
IFVD             & 0.7264          & 0.5627          & 0.4536          & 0.9716          & 0.9608 & 0.9564          & 0.9786          & 0.9351          & 0.9311          & 0.7421          & 0.6551          & 0.5947          & 0.7802          & 0.7232          & 0.6890          \\
CoCo             & 0.7323          & 0.5644          & 0.4531          & 0.9723          & 0.9586          & 0.9566          & 0.9799          & 0.9391          & 0.9237          & 0.7476          & 0.6646          & 0.5958          & 0.7840          & 0.7273          & 0.6897          \\ \hline
\textbf{GKD}     & \textcolor{blue}{\textbf{0.7804}} & \textcolor{blue}{\textbf{0.6252}} & \textcolor{blue}{\textbf{0.5061}} & \textbf{0.9728} & \textbf{0.9612}         & \textbf{0.9582} & \textbf{0.9837} & \textbf{0.9504} & \textbf{0.9490} & \textbf{0.7637} & \textbf{0.6803} & \textbf{0.6309} & \textbf{0.7940} & \textbf{0.7375} & \textbf{0.7087} \\ \bottomrule
\end{tabular}%
}
\vspace{-5mm}
\end{table*}

\begin{figure}[t]
\begin{center}
\includegraphics[width=0.9\linewidth]{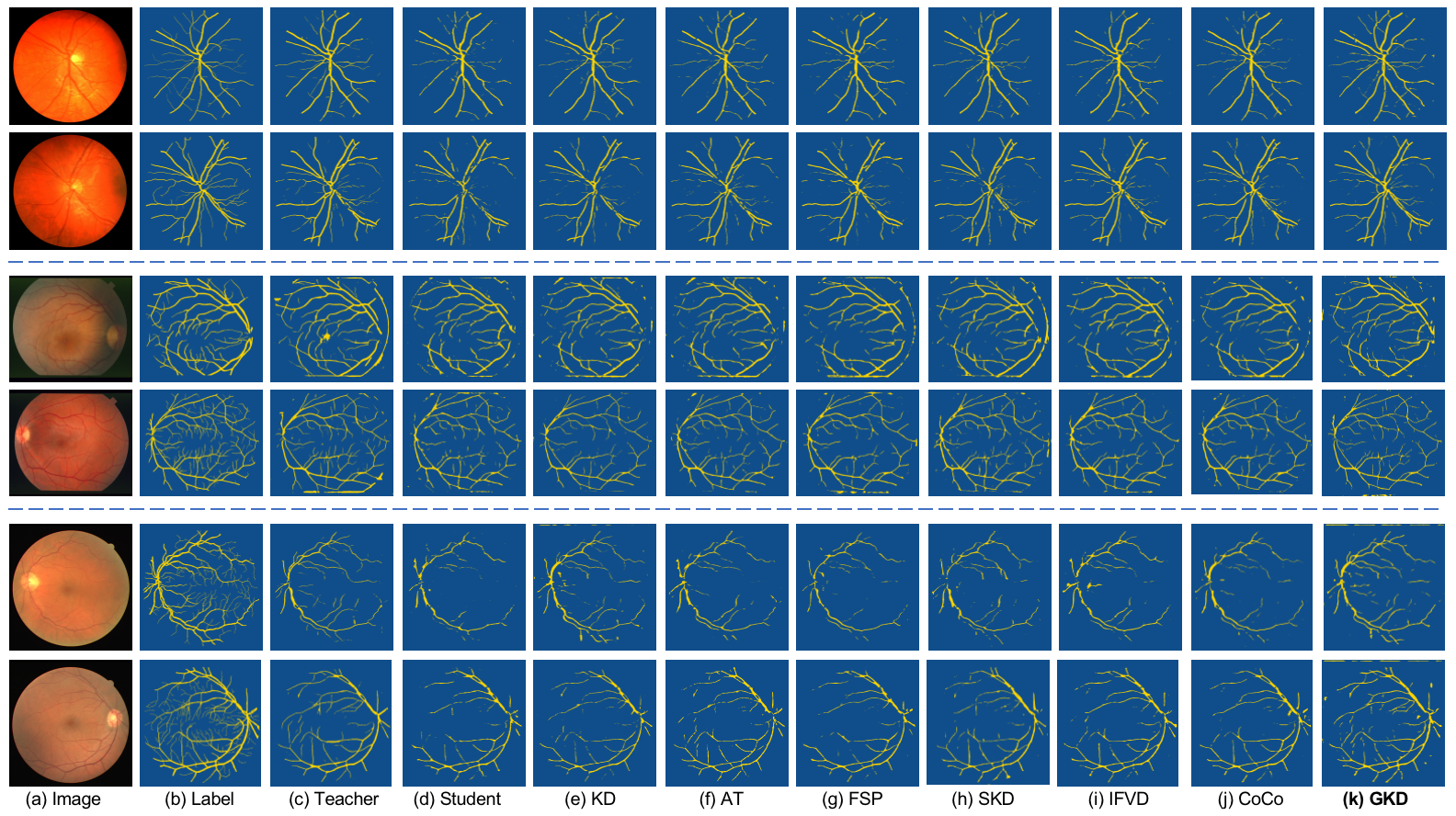}
\vspace{-4mm}
\end{center}\caption{Retinal vessel segmentation comparisons of the proposed GKD with previous knowledge distillation counterparts. 
The top two-row samples are from the CHASEDB1 dataset, the middle two-row samples are from the STARE dataset, and the bottom two-row samples are from the DRIVE dataset. \textbf{Zoom in for better details.}
}
\vspace{-6mm}
\label{fig:vessel_results}
\end{figure}
\vspace{-3mm}
\subsection{Evaluation Metrics}
\subsubsection{Metrics on Semantic Segmentation}
We adopt various metrics to evaluate the segmentation performance of the proposed GKD framework, including Sensitivity/Recall (dubbed SE), pixel Accuracy (ACC), AUC (the area under the receiving operator characteristics (ROC) curve), F1 score and mIOU. 
Since the liver and retinal vessel segmentation tasks are more focused on the foreground semantic classes (\emph{i.e.} the liver and vessel pixels), we treat the SE as the main metric. 
Additionally, we introduce the FLoating-point Operations Per second (FLOPs) and the network Parameters (Params) to measure the model complexity in our experiments.
\subsubsection{\textbf{Fréchet Semantic Distance}}
To further evaluate the generalizable ability of the student model optimized by our proposed GKD, we design a \textbf{new quantitative metric} named Fréchet Semantic Distance (FSD). 
In the aforementioned ACT, we pretrain a semantic autoencoder.
Accordingly, we argue that the semantic autoencoder can be treated as a domain-invariant semantic feature extractor. 
Inspired by the FID \cite{b46} metric, we define the FSD as the Fréchet distance between the latent features extracted from semantic autoencoder and pre-trained MSAN encoding headers, respectively.  
Specifically, the distance is formulated as Gaussian mean and covariance of both features as follows:
\begin{align}
FSD(X,Y) = \left \| \mu _{x} -\mu _{y}  \right \| +Tr( {\textstyle \sum_{x}} + {\textstyle \sum_{y}} - 2({\textstyle \sum_{x}}{\textstyle \sum_{y}})^{1/2}), 
\end{align}
where $X$ and $Y$ are the input images and the paired labels for MSAN and the semantic autoencoder, respectively. 
$\mu _{x}$ and $\mu _{y}$ are Gaussian mean of extracted features; ${\textstyle \sum_{x}}$ and ${\textstyle \sum_{y}}$ are covariance of extracted features. 
FSD estimates the effectiveness of extracting the domain-invariant semantic knowledge.

\begin{table}[t]
\centering
\caption{Ablation study on CHAOS and LITS datatsets. 
}
\label{tab:my-table3}
\footnotesize
\setlength{\tabcolsep}{1.0mm}{%
\renewcommand{\arraystretch}{0.98} 
\begin{tabular}{c|c|ccccc}
\toprule
Model & Dataset & SE$\uparrow$ & ACC$\uparrow$ & AUC$\uparrow$ & F1$\uparrow$ & mIOU$\uparrow$ \\ \hline
\multirow{3}{*}{\textbf{T: FANet}} & CHAOS & 0.9536 & 0.9928  & 0.9985 & 0.9361 & 0.9362 \\
 & LITS & 0.8251 & 0.9825  & 0.9872 & 0.8556 & 0.8646 \\ \cline{2-7} 
 & GAP$\downarrow$ & 0.1285 & 0.0103  & 0.0113 & 0.805 & 0.0716 \\ \hline
\multirow{3}{*}{S: Mobile Unet} & CHAOS & 0.8840 & 0.9899  & 0.9972 & 0.9057 & 0.9085 \\
 & LITS & 0.4812 & 0.9625  & 0.9723 & 0.6180 & 0.7042 \\ \cline{2-7} 
 & GAP$\downarrow$ & 0.4028 & 0.0274  & 0.0249 & 0.2877 & 0.2043 \\ \hline
\multirow{3}{*}{S: w/ Intra} & CHAOS & 0.9194 & 0.9911  & 0.9978 & 0.9189 & 0.9203 \\
 & LITS & 0.6831 & 0.9747  & 0.9876 & 0.7730 & 0.8018 \\ \cline{2-7} 
 & GAP$\downarrow$ & \textcolor{blue}{0.2363} & 0.0164  & 0.0102 & 0.1459 & 0.1185 \\ \hline
\multirow{3}{*}{\begin{tabular}[c]{@{}c@{}}S: w/ Intra\\ + Inter\end{tabular}} & CHAOS & \textbf{0.9474} & 0.9915  & 0.9983 & 0.9244 & 0.9252 \\
 & LITS & 0.7632 & 0.9780  & 0.9882 & 0.8145 & 0.8320 \\ \cline{2-7} 
 & GAP$\downarrow$ & \textcolor{blue}{0.1842} & 0.0135  & 0.0101 & 0.1099 & 0.0932 \\ \hline
\multirow{3}{*}{\textbf{\begin{tabular}[c]{@{}c@{}}S: w/ Intra\\ + Inter\\ + Cross\end{tabular}}} & CHAOS &  0.9451 &\textbf{0.9924}  & \textbf{0.9984} & \textbf{0.9320} & \textbf{0.9323} \\
 & LITS & \textbf{0.8251} & \textbf{0.9825}  & \textbf{0.9912} & \textbf{0.8560} & \textbf{0.8649} \\ \cline{2-7} 
 & GAP$\downarrow$ & \textbf{0.1200} & \textbf{0.0099}  & \textbf{0.0072} & \textbf{0.0760} & \textbf{0.0674} \\ \bottomrule
\end{tabular}%
}
\vspace{-4mm}
\end{table}

\vspace{-5mm}
\subsection{Implementation Details}
\vspace{-1mm}
To demonstrate the priority of our proposed GKD framework, we adopt FANet \cite{b35} as the powerful teacher model which shows SOTA performance on medical image segmentation tasks. 
Moreover, we employ the various lightweight models (\emph{i.e.} MobileNetv2 \cite{b48}, ENet \cite{b49}) as students to verify the generalization of our framework. 
We take the MobileNetv2 as the main student model in the following experiments.
All the models are implemented on the Pytorch platform with 2 NVIDIA TITAN XP GPUs. 
The batchsize is set to 16 for both teacher and student networks. 
In the training phase, we adopt the initial learning rate as 0.003 and the total optimization epoch is 100.

\begin{figure}[t]
\begin{center}
\includegraphics[width=0.85\linewidth]{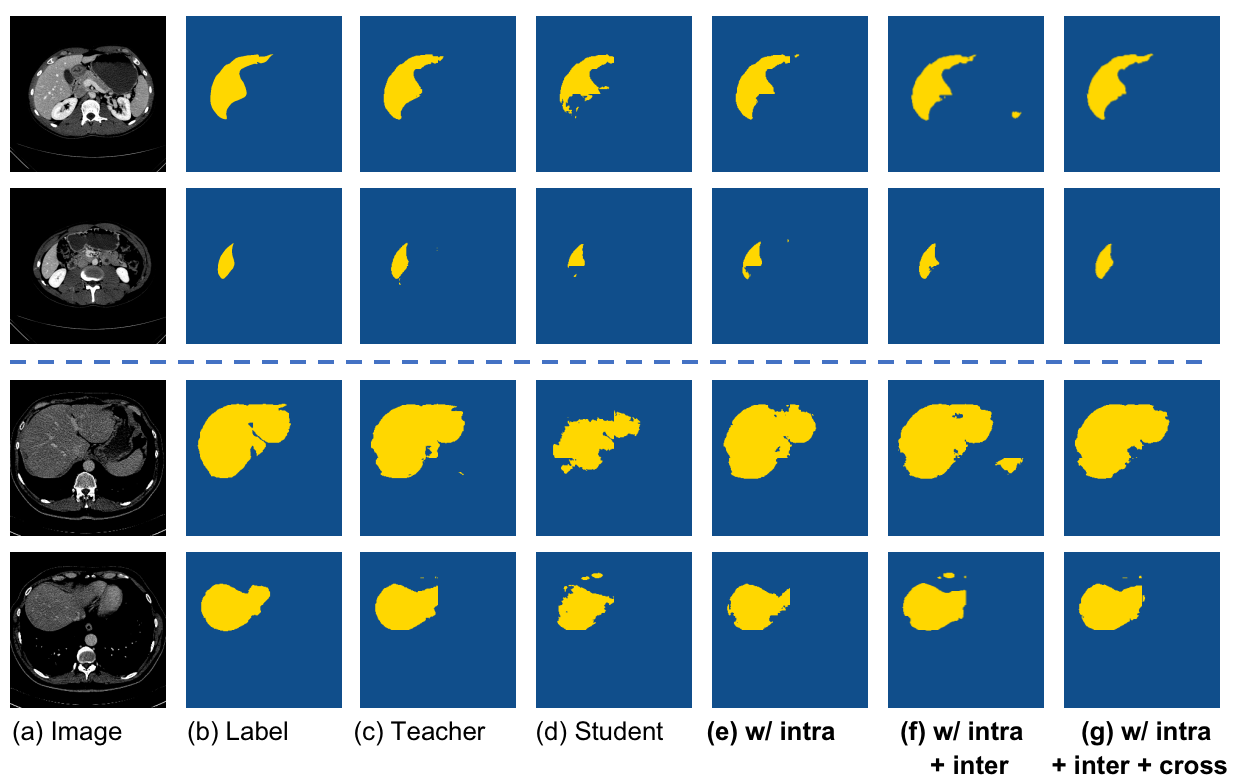}
\vspace{-6mm}
\end{center}\caption{The visual comparisons of ablation study on the liver segmentation task. The top two-row samples are from CHAOS dataset, and the bottom two-row samples are from the LITS dataset. 
}
\vspace{-5mm}
\label{fig:liver_abliation}
\end{figure}

\begin{table}[t]
\centering
\caption{Ablation study on CHASEDB1, STARE and DRIVE datasets.
}
\label{tab:my-table4}
\footnotesize
\setlength{\tabcolsep}{1.0mm}{%
\renewcommand{\arraystretch}{1.0} 
\begin{tabular}{c|c|ccccc}
\toprule
Model & Dataset & SE$\uparrow$ & ACC$\uparrow$ & AUC$\uparrow$ & F1$\uparrow$ & mIOU$\uparrow$ \\ \hline
\multirow{3}{*}{\textbf{T: FANet}} & CHASEDE1 & \multicolumn{1}{l}{0.8102} & \multicolumn{1}{l}{0.9730} & \multicolumn{1}{l}{0.9836} & \multicolumn{1}{l}{0.7711} & \multicolumn{1}{l}{0.7996} \\
 & STARE & 0.6485 & 0.9544 & 0.9541 & 0.6526 & 0.7183 \\
 & DRIVE & 0.5444 & 0.9697 & 0.9561 & 0.6558 & 0.7229 \\ \hline
\multirow{3}{*}{S: Mobile Unet} & CHASEDE1 & 0.6894 & 0.9686 & 0.9737 & 0.7113 & 0.7596 \\
 & STARE & 0.5050 & 0.9575 & 0.9244 & 0.6106 & 0.6977 \\
 & DRIVE & 0.4174 & 0.9543 & 0.9320 & 0.5630 & 0.6723 \\ \hline
\multirow{3}{*}{S: w/ intra} & CHASEDE1 & 0.7437 & 0.9700 & 0.9777 & 0.7354 & 0.7751 \\
 & STARE & 0.5528 & 0.9586 & 0.9340 & 0.6381 & 0.7127 \\
 & DRIVE & \textcolor{blue}{0.4493} & 0.9490 & 0.9242 & 0.5542 & 0.6653 \\ \hline
\multirow{3}{*}{\begin{tabular}[c]{@{}c@{}}S: w/ intra\\ +inter\end{tabular}} & CHASEDE1 & 0.7559 & 0.9710 & 0.9795 & 0.7453 & 0.7818 \\
 & STARE & 0.6138 & 0.9583 & 0.9361 & 0.6605 & 0.7248 \\
 & DRIVE & \textcolor{blue}{0.4974} & 0.9557 & 0.9314 & 0.6130 & 0.6980 \\ \hline
\multirow{3}{*}{\textbf{\begin{tabular}[c]{@{}c@{}}S: w/ intra\\ + inter\\ + cross\end{tabular}}} & CHASEDE1 & \textbf{0.7804} & \textbf{0.9728} & \textbf{0.9837} & \textbf{0.7637} & \textbf{0.7940} \\
 & STARE & \textbf{0.6252} & \textbf{0.9612} & \textbf{0.9504} & \textbf{0.6803} & \textbf{0.7375} \\
 & DRIVE & \textbf{0.5061} & \textbf{0.9582} & \textbf{0.9490} & \textbf{0.6309} & \textbf{0.7087} \\  \bottomrule
\end{tabular}%
}
\vspace{-6mm}
\end{table}

\vspace{-5mm}
\subsection{Comparative Experiments}
\vspace{-1mm}
\subsubsection{Experiments on Liver Segmentation} We compare our method with various knowledge distillation approaches: KD \cite{b18}, AT \cite{b19}, FSP \cite{b20}, SKD \cite{b21}, IFVD \cite{b22}, CoCo \cite{b23}. 
As reported in Tab.~\ref{tab:my-table1}, our proposed GKD outperforms these methods with a large margin on liver segmentation tasks. 
All the models in Tab.~\ref{tab:my-table1} are trained only on the CHAOS dataset. 
We directly test the models on the LITS dataset to demonstrate the priority of both high performance and generalizable ability. Specifically, to evaluate the generalizable ability of our GKD, we calculate the performance \textbf{GAP} between these two datasets. 
As for the teacher model, due to its complicated network architecture, there is a slight drop when testing directly on the LITS dataset.
However, the original shallow student model suffers from a big performance drop when tested on the LITS dataset. Although the previous distillation methods obtain apparent performance improvement in the CHAOS testing phase. 
They still fail at generalization testing on the LITS dataset. 
In contrast, our proposed GKD allows the student network to achieve competitive performance with teacher model on AUC evaluation indicator in LITS testing phase (\emph{i.e.} GKD: Student 0.9912 $vs$ Teacher 0.9908). 
Remarkably, as for \textbf{GAP} between the testing phase on CHAOS and LITS, our GKD even reach better results than the teacher model with several evaluation indicators (\emph{e.g.} SE: GKD 0.1200 $vs$ Teacher 0.1257)
The narrowed GAP fully verifies the effectiveness of our proposed method for improving the generalization ability of the student network.

Additionally, from the perspective of qualitative evaluation, we show the visual comparison results predicted by various distillation methods and our GKD in Fig.~\ref{fig:liver_results}. Obviously, the image structures and textures of the CHAOS and the LITS datasets are quite different. While the other distillation methods fail to segment the liver foreground and distinguish background pixels correctly, our method consistently surpasses the counterparts on cross-domain scenes.

\begin{figure}[t]
\begin{center}
\includegraphics[width=0.85\linewidth]{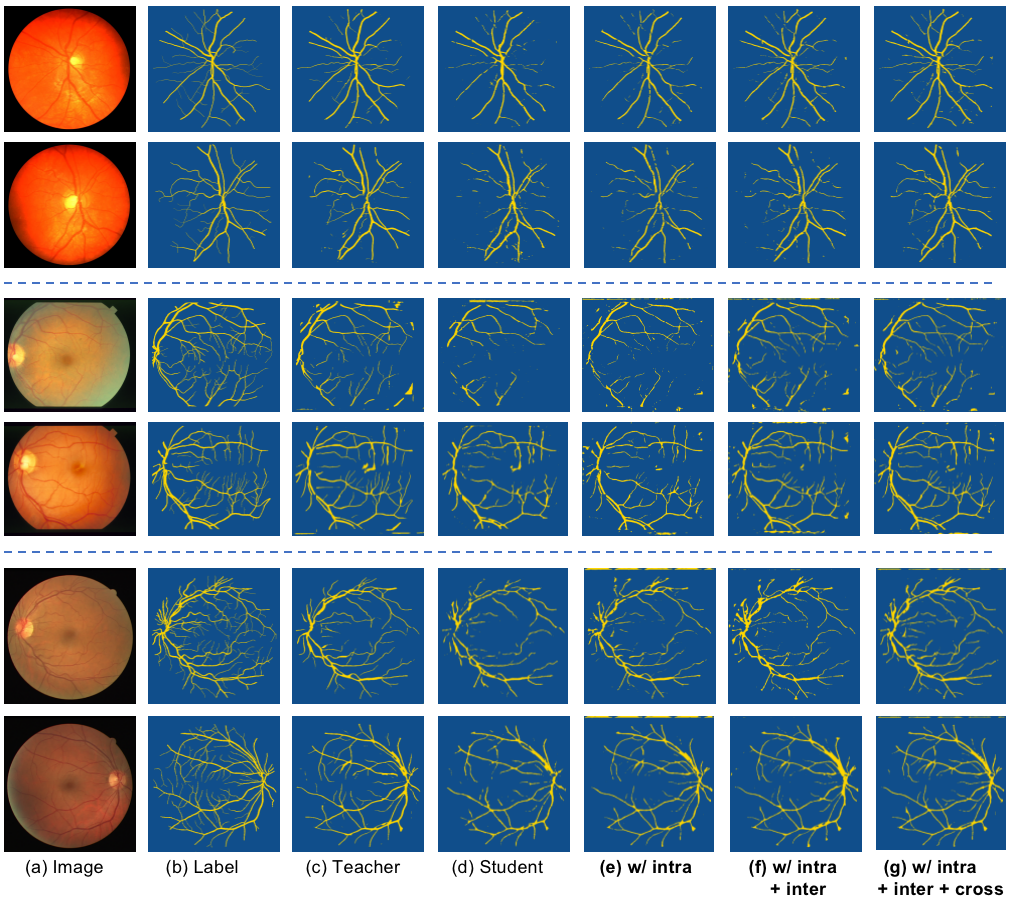}
\vspace{-6mm}
\end{center}\caption{The visual comparisons of ablation study on the retinal vessel segmentation task. The top two-row samples are from CHASEDB1 dataset, the middel two-row samples are from STARE dataset, and the bottom two-row samples are from the DRIVE dataset. 
}
\vspace{-5mm}
\label{fig:vessel_abliation}
\end{figure}

\begin{table}[t]
\centering
\caption{FSD results on retinal vessel and liver segmentation tasks. 
}
\label{tab:my-table5}
\footnotesize
\setlength{\tabcolsep}{1.0mm}{%
\renewcommand{\arraystretch}{0.88} 
\begin{tabular}{c|ccccc}
\toprule
Model & \multicolumn{5}{c}{Fréchet Semantic Distance (FSD) $\downarrow$} \\ \hline
Dataset & CHASEDB1 & STARE & \multicolumn{1}{c|}{DRIVE} & CHAOS & LITS \\ \hline
\begin{tabular}[c]{@{}c@{}}T: FANet\\ w/ o\end{tabular} & 53.28 & 55.94 & \multicolumn{1}{c|}{56.31} & 79.93 & 82.29 \\
\begin{tabular}[c]{@{}c@{}}S:Mobile Unet\\  w/ o\end{tabular} & 1660.94 & 2331.28 & \multicolumn{1}{c|}{2790.26} & 1445.79 & 1682.39 \\ \hline
\begin{tabular}[c]{@{}c@{}}T: FANet\\ w/ A\end{tabular} & 34.23 & 35.40 & \multicolumn{1}{c|}{36.29} & 45.73 & 46.84 \\
\begin{tabular}[c]{@{}c@{}}S: Mobile Unet\\ w/ A\end{tabular} & 146.85 & 147.17 & \multicolumn{1}{c|}{148.31} & 121.52 & 123.00 \\ \hline
\textbf{S: GKD} & \textbf{55.11} & \textbf{55.51} & \multicolumn{1}{c|}{\textbf{56.06}} & \textbf{45.00} & \textbf{49.37} \\ \bottomrule
\end{tabular}%
}
\vspace{-6mm}
\end{table}

\subsubsection{Experiments on Retinal Vessel Segmentation}

As reported in Tab.~\ref{tab:my-table2}, we further conduct comparative experiments on retinal vessel segmentation tasks. 
In this phase, we introduce three datasets to verify the high segmentation performance and generalizable ability of our proposed GKD. 
All the models involved in these experiments are trained on the CHASEDB1 training set and directly tested on the three vessel segmentation datasets. 
Our method achieves superior performance compared with other distillation ones. 
Quantitatively, as for the SE indicator, the proposed GKD outperforms the trained-from-scratch student model by a large margin of 13.2\% (\emph{i.e.} $(0.7804-0.6894) / 0.6894 \approx 13.2\%$), colored in \textcolor{blue}{blue} for better view.) on the CHASEDB1 dataset. Moreover, as for the testing phase on STARE and DRIVE, the GKD improves the SE by about 23.8\% and 21.3\% compared with the trained-from-scratch student. Besides, our GKD even achieves better results than the teacher model with several evaluation indicators such as ACC, F1 score, and mIOU in the STARE testing phase (\emph{e.g.} F1: 0.6803 $vs$ 0.6526). 

Qualitatively, we illustrate the visual comparison results segmented by other distillation counterparts and our GKD in Fig.~\ref{fig:vessel_results}. 
It is obvious that there are domain gaps (\emph{e.g.} illumination) among the three datasets. 
To be specific, the retinal vessel images in the CHASEDB1 dataset are much more bright-some than another two datasets. 
In addition, the different shooting angles also lead to variability in vessel structure and texture. 
The aforementioned domain-gap results in a large margin decline of the existing methods when directly testing on STARE and DRIVE. 
On the contrary, the last column in Fig.~\ref{fig:vessel_results} demonstrates that our method has a stronger generalizable ability than previous ones. 

\begin{table*}[t]
\centering
\caption{Results on CHAOS and LITS datasets with different students. $\uparrow$means the higher, the better. $\downarrow$means the lower, the better.}
\label{tab:my-table6}
\footnotesize
\resizebox{\textwidth}{!}{%
\renewcommand{\arraystretch}{1.0} 
\begin{tabular}{cccc|ccc|ccc|ccc|ccc|ccc}
\toprule
\multicolumn{4}{c|}{Model} & \multicolumn{3}{c|}{SE $\uparrow$} & \multicolumn{3}{c|}{ACC $\uparrow$} & \multicolumn{3}{c|}{AUC $\uparrow$} & \multicolumn{3}{c|}{F1 $\uparrow$} & \multicolumn{3}{c}{mIOU $\uparrow$} \\ \hline
\multicolumn{2}{c|}{Datasets} & \textbf{FLOPs(G) $\downarrow$} & \textbf{Parms(M) $\downarrow$} & CHAOS & \multicolumn{1}{c|}{LITS} & \textbf{GAP$\downarrow$} & CHAOS & \multicolumn{1}{c|}{LITS} & \textbf{GAP$\downarrow$} & CHAOS & \multicolumn{1}{c|}{LITS} & \textbf{GAP$\downarrow$} & CHAOS & \multicolumn{1}{c|}{LITS} & \textbf{GAP$\downarrow$} & CHAOS & \multicolumn{1}{c|}{LITS} & \textbf{GAP$\downarrow$} \\ \hline
\multicolumn{1}{c|}{Teacher} & \multicolumn{1}{c|}{FANet} & 171.556 & 38.250 & 0.9613 & \multicolumn{1}{c|}{0.8356} & 0.1257 & 0.9932 & \multicolumn{1}{c|}{0.9835} & 0.0097 & 0.9986 & \multicolumn{1}{c|}{0.9908} & 0.0078 & 0.9401 & \multicolumn{1}{c|}{0.8647} & 0.0754 & 0.9399 & \multicolumn{1}{c|}{0.8721} & 0.0678 \\ \hline
\multicolumn{1}{c|}{\multirow{7}{*}{Student}} & \multicolumn{1}{c|}{\begin{tabular}[c]{@{}c@{}}Mobile Unet \\ w/ o\end{tabular}} & \multirow{3}{*}{1.492} & \multirow{3}{*}{4.640} & 0.8840 & \multicolumn{1}{c|}{0.4812} & 0.4028 & 0.9899 & \multicolumn{1}{c|}{0.9625} & 0.0274 & 0.9972 & \multicolumn{1}{c|}{0.9723} & 0.0249 & 0.9057 & \multicolumn{1}{c|}{0.6180} & 0.2877 & 0.9085 & \multicolumn{1}{c|}{0.7042} & 0.2043 \\
\multicolumn{1}{c|}{} & \multicolumn{1}{c|}{\textbf{\begin{tabular}[c]{@{}c@{}}Mobile Unet\\ w/ GKD\end{tabular}}} &  &  & \textbf{0.9451} & \multicolumn{1}{c|}{\textbf{0.8251}} & \textbf{0.1200} & \textbf{0.9924} & \multicolumn{1}{c|}{\textbf{0.9825}} & \textbf{0.0099} & \textbf{0.9984} & \multicolumn{1}{c|}{\textbf{0.9912}} & \textbf{0.0072} & \textbf{0.9320} & \multicolumn{1}{c|}{\textbf{0.8560}} & \textbf{0.0760} & \textbf{0.9323} & \multicolumn{1}{c|}{\textbf{0.8649}} & \textbf{0.0674} \\ \cline{2-19} 
\multicolumn{1}{c|}{} & \multicolumn{1}{c|}{\begin{tabular}[c]{@{}c@{}}ENet\\ w/ o\end{tabular}} & \multirow{3}{*}{0.516} & \multirow{3}{*}{0.349} & 0.8924 & \multicolumn{1}{c|}{0.6816} & 0.2108 & 0.9851 & \multicolumn{1}{c|}{0.9175} & 0.0136 & 0.9940 & \multicolumn{1}{c|}{0.9555} & 0.0385 & 0.8682 & \multicolumn{1}{c|}{0.7510} & 0.1172 & 0.9757 & \multicolumn{1}{c|}{0.7857} & 0.1900 \\
\multicolumn{1}{c|}{} & \multicolumn{1}{c|}{\textbf{\begin{tabular}[c]{@{}c@{}}ENet\\ w/ GKD\end{tabular}}} &  &  & \textbf{0.9550} & \multicolumn{1}{c|}{\textbf{0.8326}} & \textbf{0.1224} & \textbf{0.9919} & \multicolumn{1}{c|}{\textbf{0.9807}} & \textbf{0.0112} & \textbf{0.9974} & \multicolumn{1}{c|}{\textbf{0.9811}} & \textbf{0.0173} & \textbf{0.9281} & \multicolumn{1}{c|}{\textbf{0.8445}} & \textbf{0.0836} & \textbf{0.9287} & \multicolumn{1}{c|}{\textbf{0.8552}} & \textbf{0.0735} \\ \bottomrule
\end{tabular}%
}
\vspace{-6mm}
\end{table*}

\vspace{-3mm}
\subsection{Ablation Study}
\vspace{-1mm}
In this subsection, we conduct the ablation study on the distillation constraints of our GKD. 
For brevity, we report the experimental results over 5 testing datasets, simultaneously.

\subsubsection{Ablation on Intra-coupling Graph Distillation}
Quantitatively, we analyze the impact of the proposed Intra-coupling Graph Distillation schema on both liver segmentation and retinal vessel segmentation tasks. 
Our intra-coupling graph demonstrates the semantic-shift between the anchor features and features of each augmented sample with different couplings, caused by data augmentation. 
By enforcing the student to mimic representative vectors of the teacher in the regularized latent space, the segmentation performance and generalization ability of student are significantly improved. 
As shown in Tab.~\ref{tab:my-table3}, the middle three-row shows that our intra-coupling distillation could provide generalizable knowledge for student. 
Concretely, the performance gap is significantly decreasing from 0.4028 to 0.2363 on the SE indicator. 
Meanwhile, performance improves significantly with all metrics, compared with the trained-from-scratch student in the CHAOS testing phase. 
Moreover, when intra-coupling distillation is utilized, the student performance on the three retinal vessel segmentation datasets is significantly improved. 
Qualitatively, we visualize the results of the above ablation experiments in Fig.~\ref{fig:liver_abliation} and Fig.~\ref{fig:vessel_abliation}. 
After adding intra-coupling graph supervision, the student model is more sensitive to details.

\subsubsection{Ablation on Inter-coupling Graph Distillation}
To evaluate the influence of our proposed Inter-coupling Graph Distillation schema, we conduct experiments by accumulating this supervision on the previous versions as reported in Tab.~\ref{tab:my-table3} and Tab.~\ref{tab:my-table4}. 
As for the SE indicator, the student model achieves the best performance of 0.9474 in the CHAOS dataset. 
Meanwhile, the gap between the two liver segmentation datasets is obviously reduced. 
Specifically, there is a 22\% decrease compared to only leveraging the intra-coupling graph distillation schema. 
In the retina vessel segmentation testing phase, all the metrics reach higher tiers compared with the last model version. 
Concretely, the SE indicator achieves about 10.7\% improvement on the DRIVE dataset. 
This indicates the essence of inter-coupling graph distillation on cross-domain medical image segmentation.
After adopting inter-coupling graph distillation, the incorrect segmentation of indistinguishable background pixels is greatly suppressed, as could be observed from results under  `w/ intra + inter' in Fig.~\ref{fig:liver_abliation} and Fig.~\ref{fig:vessel_abliation}.

\subsubsection{Ablation on Domain-Invariant Cross Distillation}
We add the Domain-Invariant Cross Distillation (DICD) schema as the final component of our proposed GKD framework. In DICD, the model-specific features are cross-reconstructed by exchanging the decoding headers of MSAN. 
The features reconstructed from regularized domain-invariant vectors contain more pure semantic information intuitively, as shown in Fig.~\ref{fig:ACT}. 
Such semantic information facilitates the generalization of the student model in cross-domain scenes.
We leverage the powerful pre-trained teacher model to provide the supervision of the encoder and decoder of student jointly. Thus, after adding the DICD, the whole performance of the student model is further improved. 
Segmented results obtained by different variants of the proposed model are shown in Fig.~\ref{fig:liver_abliation} and Fig.~\ref{fig:vessel_abliation}. It could be observed that the pixel prediction results of the semantic foreground are gradually optimized via a scratch-trained model. Although in the final version, the SE metric on the CHAOS dataset slightly declined, the GKD achieves the best results
in most metrics as reported in Tab.~\ref{tab:my-table3} and Tab.~\ref{tab:my-table4}.

\vspace{-3mm}
\subsection{Domain-invariant Verification of GKD}
\vspace{-1mm}
To further verify the domain-invariant representation ability of our proposed MSAN, we calculate the FSD between the latent vectors from SAN (or TAN) and the pre-trained semantic autoencoder.
In ACT, the main step is to provide the domain-invariant regularization (the word `Alignment' in Fig.~\ref{fig:ACT}) for the TAN and SAN. 
This could restrict the latent vectors from these two models to align in the consistent semantic space, leading to domain-invariant representations. 
As reported in Tab.~\ref{tab:my-table5}, 
`w/ o' denotes dropping the alignment step when training the MSAN; `w/ A' denotes adopting the alignment step. 
After the alignment operation, the FSD of the scratch-trained student model is dramatically decreased, which means the domain-specific features are mapped to be more domain-invariant.
This also indicates the latent vectors embedded by SAN are more consistent with vectors embedded by TAN. 
Furthermore, after conducting the DCGD and DICD schemes, our GKD achieves much lower FSD than the scratch-trained student model with the 'w/ A' operation (\emph{i.e.} $1660.94\to146.85\to55.11$ in CHASEDB1 dataset).
This proves that our GKD further facilitates better domain-invariant representations progressively. As for the direct testing datasets (\emph{i.e.} STARE, DRIVE, and LITS), the FSD of the student model is also significantly decreased after conducting our proposed GKD.
\vspace{-3mm}
\subsection{Generalization on Different Students}
\vspace{-1mm}
To further evaluate the generalization of our proposed GKD with different shallow network architectures, we adopt the lightweight ENet \cite{b49} as an alternative student. 
The comparison results are reported in Tab.~\ref{tab:my-table6}, where all the hyper-parameters are consistent with previous experiments.
We observed that both Mobile Unet based and ENet based students have much lower FLOPs and Params than the complicated teacher model. 
Their shallow network architectures result in poor performance on the CHAOS and LITS datasets. 
However, after the distillation by GKD, both lightweight student models obtain significant improvements in all metrics.

\section{Conclusion}
\vspace{-1mm}
In this paper, we propose a generalizable knowledge distillation framework for efficient cross-domain medical image segmentation, namely GKD. 
In GKD, we take full advantage of a pre-trained semantic autoencoder to provide the domain-invariant regularization for training the Model-Specific Alignment Networks (MSAN). 
Guided by MSAN, two generalizable knowledge distillation schemes named Dual Contrastive Graph Distillation and Domain-Invariant Cross Distillation are proposed to improve the performance of lightweight models in cross-domain scenes. 
Moreover, we introduce a new metric named Fréchet Semantic Distance to verify the effectiveness of the regularized domain-invariant features. 
As a result, our method achieves outstanding performance compared with various knowledge distillation counterparts on the liver and retinal vessel segmentation tasks. 
In the future, we would like to investigate the knowledge distillation approaches for cross-modality medical image analysis.


 






\vspace{-4mm}
\begin{thebibliography}{00}

\bibitem{b1} O. Ronneberger, et. al, "U-net: Convolutional networks for biomedical image segmentation," in \emph{Proc. Int. Conf. Med. Image Comput. Comput. Assist. Intervent. (MICCAI)}, 2015, pp. 234–241.

\bibitem{b2} Z. Zhou,et. al, "UNet++: Redesigning skip connections to exploit multiscale features in image segmentation, " \emph{IEEE Trans. Med. Imag. (TMI)}, vol. 39, no. 6, pp. 1856–1867, 2019. 

\bibitem{b3} F. Milletari, et. al, "V-net: Fully convolutional neural networks for volumetric medical image segmentation," In \emph{Proc.Fourth Int. Conf. 3D Vis. (3DV)}, Oct. 2016, pp. 565–571.

\bibitem{b4} Y. Jiang, et. al, "A Novel Negative-Transfer-Resistant Fuzzy Clustering Model With a Shared Cross-Domain Transfer Latent Space and its Application to Brain CT Image Segmentation,"  \emph{IEEE/ACM Trans. Comput. Biol. Bioinf.}, vol. 18, no. 1, pp. 40-52, Feb. 2021.

\bibitem{b5} X. Liu, et. al, "Adapting off-the-shelf source segmenter for target medical image segmentation," in \emph{Proc. Int. Conf. Med. Image Comput. Comput. Assist. Intervent. (MICCAI)}, Sept. 2021, pp. 549-559.

\bibitem{b6} D. Zou, et. al, “Unsupervised domain adaptation with dualscheme fusion network for medical image segmentation,” in \emph{Proc. 29th Int. Joint Conf. Artif. Intell. (IJCAI),} 2020, pp. 3291–3298

\bibitem{b7} X. Han et al., "Deep Symmetric Adaptation Network for Cross-Modality Medical Image Segmentation," \emph{IEEE Trans. Med. Imag. (TMI)}, vol. 41, no. 1, pp. 121-132, Jan. 2022. 

\bibitem{b8} X. Han, et. al, "Anatomy-Regularized Representation Learning for Cross-Modality Medical Image Segmentation," \emph{IEEE Trans. Med. Imag. (TMI)}, vol. 40, no. 1, pp. 274-285, Jan. 2021.


\bibitem{b10} M. Perslev, et. al, "One network to segment them all: A general, lightweight system for accurate 3d medical image segmentation," in \emph{Proc. Int. Conf. Med. Image Comput. Comput. Assist. Intervent. (MICCAI)} Oct. 2019, pp. 30-38.

\bibitem{b11} D. Qin, et. al, "Efficient Medical Image Segmentation Based on Knowledge Distillation," \emph{IEEE Trans. Med. Imag. (TMI)}, vol. 40, no. 12, pp. 3820-3831, Dec. 2021.

\bibitem{b12} A. G. Howard, et. al, "Mobilenets: Efficient convolutional neural networks for mobile vision applications," 2017, \emph{arXiv:1704.04861.} [Online]. Available: https://arxiv.org/abs/1704.04861.


\bibitem{b14} X. Li, et. al, "Selective kernel networks, " in \emph{Proc. IEEE/CVF Conf. Comput. Vis. Pattern Recognit. (CVPR)}, Jun. 2019, pp. 510-519.

\bibitem{b15} Y. He, et. al, "Filter pruning via geometric median for deep convolutional neural networks acceleration," in \emph{Proc. IEEE/CVF Conf. Comput. Vis. Pattern Recognit. (CVPR)}, Jun. 2019, pp. 4340-4349.


\bibitem{b17} Z. Wang, et. al, "Convolutional neural network pruning with structural redundancy reduction", in \emph{Proc. IEEE/CVF Conf. Comput. Vis. Pattern Recognit. (CVPR)}, Jun. 2021, pp. 14913-14922.

\bibitem{b18} G. Hinton, et. al, "Distilling the knowledge in a neural network.," 2015, \emph{arXiv:1503.02531.} [Online]. Available: https://arxiv.org/abs/1503.02531.

\bibitem{b19} S. Zagoruyko, et. al, "Paying more attention to attention: improving the performance of convolutional neural networks via attention transfer,"  in \emph{Proc. Int. Conf. Learn. Represent. (ICLR),}. Jun. 2017, pp. 1-13.

\bibitem{b20} J. Yim, D. Joo, et. al, "A gift from knowledge distillation: Fast optimization, network minimization and transfer learning," in \emph{Proc. IEEE Conf. Comput. Vis. Pattern Recognit. (CVPR),} Jul. 2017, pp. 4133–4141.

\bibitem{b21} Y. Liu, et. al, "Structured knowledge distillation for dense prediction," \emph{IEEE Trans. Pattern Anal. Mach. Intell. (TPAMI),} Jun. 2020.

\bibitem{b22} Y. Wang, et. al, "Intra-class feature variation distillation for semantic segmentation," in \emph{Proc. Euro. Conf. Comput. Vis. (ECCV),} Nov. 2020, pp. 346–362.

\bibitem{b23} W. Zou, et. al, "CoCo distillnet: a cross-layer correlation distillation network for pathological gastric cancer segmentation," in \emph{Proc. IEEE Int. Conf. Bioinform. Biomed. (BIBM),} Dec. 2021, pp. 1227–1234.


\bibitem{b25} Y. Zhou, et. al, "Deep semi-supervised knowledge distillation for overlapping cervical cell instance segmentation.," in \emph{Proc. Int. Conf. Med. Image Comput. Comput. Assist. Intervent. (MICCAI)}, Oct. 2020, pp. 521-531.

\bibitem{b26} R. Aljundi, et. al, "Lightweight unsupervised domain adaptation by convolutional filter reconstruction," in \emph{Proc. Eur. Conf. Comput. Vis. (ECCV)}, Oct. 2016, pp. 508-515. 

\bibitem{b27} S. Ye, et. al, "Light-weight calibrator: a separable component for unsupervised domain adaptation." in \emph{Proc. IEEE/CVF Conf. Comput. Vis. Pattern Recognit. (CVPR)}, 2020, pp. 13736-13745.

\bibitem{b28} S. Li, et. al, "Transferable semantic augmentation for domain adaptation," in \emph{Proc. IEEE/CVF Conf. Comput. Vis. Pattern Recognit. (CVPR)}, Jun. 2021, pp. 11516-11525.

\bibitem{b29} T. Chen, et. al, "A simple framework for contrastive learning of visual representations," in \emph{Proc. Int. Conf. Mach. Learn. (ICML),} Nov. 2020, pp. 1597-1607.

\bibitem{b30} A. V. D. Oord, et. al, "Representation learning with contrastive predictive coding," 2018, \emph{arXiv:1807.03748.} [Online]. Available: https://arxiv.org/abs/1807.03748.

\bibitem{b31} K. He, et. al, "Momentum contrast for unsupervised visual representation learning," in \emph{Proc. IEEE/CVF Conf. Comput. Vis. Pattern Recognit. (CVPR)}, Jun. 2020, pp. 9729-9738. 

\bibitem{b32} Y. Tian, et. al, "Contrastive Representation Distillation,"  in \emph{Proc. Int. Conf. Learn. Represent. (ICLR),} Sep. 2019.

\bibitem{b33} L. Wang, et. al, "Improving weakly supervised visual grounding by contrastive knowledge distillation," in \emph{Proc. IEEE/CVF Conf. Comput. Vis. Pattern Recognit. (CVPR)}, Jun. 2021, pp. 14090-14100.

\bibitem{b34} J. M. J. Valanarasu, et. al, "Kiu-net: Towards accurate segmentation of biomedical images using over-complete representations," in \emph{Proc. Int. Conf. Med. Image Comput. Comput. Assist. Intervent. (MICCAI)}, Oct. 2020, pp. 363-373.

\bibitem{b35} K. Li, et. al, "Accurate Retinal Vessel Segmentation in Color Fundus Images via Fully Attention-Based Networks," \emph{IEEE J. Biomed. Health Inform. (JBHI),} vol. 25, no. 6, pp. 2071-2081, Jun. 2021.

\bibitem{b36} S. Takahama, et. al, "Multi-stage pathological image classification using semantic segmentation," in \emph{Proc. IEEE Int. Conf. Comput. Vis. (ICCV),} Oct. 2019, pp. 10702–10711

\bibitem{b37} X. Wang, et. al, "Bix-nas: Searching efficient bi-directional architecture for medical image segmentation," in \emph{Proc. Int. Conf. Med. Image Comput. Comput. Assist. Intervent. (MICCAI)}, Sep. 2021, pp. 229-238.

\bibitem{b38} E. Creager, et. al, "Environment inference for invariant learning," in \emph{Proc. Int. Conf. Mach. Learn. (ICML),} Jul. 2021, pp. 2189-2200.


\bibitem{b40} A. E. Kavur, et. al, "CHAOS challenge-combined (CT-MR) healthy abdominal organ segmentation." \emph{Med. Image Anal.}, vol. 69, 2021.

\bibitem{b41} P. Bilic, et. al, "The liver tumor segmentation benchmark (lits)," 2019, \emph{arXiv:1901.04056.} [Online]. Available: https://arxiv.org/abs/1901.04056.

\bibitem{b42} J. Staal, et. al, "Ridge-based vessel segmentation in color images of the retina," \emph{IEEE Trans. Med. Imag. (TMI)}, vol. 23, no. 4, pp. 501-509, 2004.

\bibitem{b43} A. D. Hoover, et. al, "Locating blood vessels in retinal images by piecewise threshold probing of a matched filter response," \emph{IEEE Trans. Med. Imag. (TMI)}, vol. 19, no. 3, pp. 203-210, Mar. 2000.

\bibitem{b44} M. M. Fraz, et. al, "An ensemble classification-based approach applied to retinal blood vessel segmentation," \emph{IEEE Trans. Biomed. Eng. (TBE)}, vol. 59, no. 9, pp. 2538-2548. Sep. 2012.


\bibitem{b46} M. Lucic, et. al, “Are GANs created equal? A large-scale study,” 2017, \emph{arXiv:1706.08500,} [Online]. Available: https://arxiv.org/abs/1711.10337.


\bibitem{b48} M. Sandler, et. al, "Mobilenetv2: Inverted residuals and linear bottlenecks," in \emph{Proc. IEEE/CVF Conf. Comput. Vis. Pattern Recognit. (CVPR)}, Jun. 2018, pp. 4510-4520.

\bibitem{b49} A. Paszke, et. al, "Enet: A deep neural network architecture for real-time semantic segmentation," 2016, \emph{arXiv: 1606.02147,} [Online]. Available: https://arxiv.org/abs/1606.02147.


\bibitem{b51}
J. N. Kundu, et. al, "Non-local latent relation distillation for self-adaptive 3D human pose estimation," in \emph{Proc. Adv. Neural Inf. Process. Syst. (NIPS)}, 2021, pp. 158-171.

\bibitem{b52}
Bai, Y. et. al, "RSA: Reducing Semantic Shift from Aggressive Augmentations for Self-supervised Learning.," in \emph{Proc. Adv. Neural Inf. Process. Syst. (NIPS)}, 2022.

\end{thebibliography}
\end{document}